\documentclass{article}
\usepackage{iclr2027_conference,times}
\iclrfinalcopy
\usepackage{amsmath,amssymb,amsthm,mathrsfs}
\usepackage{graphicx,booktabs,multirow,array,wrapfig,etoolbox,float}
\usepackage{placeins}
\usepackage{url}
\usepackage{hyperref}
\AtBeginEnvironment{table}{\setlength{\abovecaptionskip}{0pt}}
\AtBeginEnvironment{wraptable}{\setlength{\abovecaptionskip}{0pt}}

\newtheorem{theorem}{Theorem}[section]
\newtheorem{proposition}[theorem]{Proposition}
\newtheorem{lemma}[theorem]{Lemma}
\newtheorem{corollary}[theorem]{Corollary}
\newcommand{\est}[2]{\shortstack[c]{#1\\$\pm$ #2}}
\newcommand{\appsection}[1]{\FloatBarrier\section{#1}\addcontentsline{atc}{section}{\protect\numberline{\thesection}#1}}
\newcommand{\appsubsection}[1]{\subsection{#1}\addcontentsline{atc}{subsection}{\protect\numberline{\thesubsection}#1}}
\newcommand{\appcontentsgroup}[1]{%
  \par\addvspace{0.45em}%
  \noindent\textsc{#1}\par\nobreak\addvspace{0.08em}%
}
\hypersetup{pdftitle={Rethinking the Verification of Machine Unlearning},pdfauthor={Lei Zhou, Youwen Zhu, Rongke Liu},pdfsubject={},pdfkeywords={}}
\title{Rethinking the Verification of\\Machine Unlearning}
\author{Lei Zhou \And Youwen Zhu \And Rongke Liu}
\begin{document}
\maketitle
\pagestyle{plain}
\begin{abstract}
Machine unlearning claims require evidence that a model has removed the
influence of specified training data. Yet widely deployable post hoc verifiers
infer unlearning largely from predictive behavior, the final model response
that remains under the provider's control. We make this limitation concrete
through Output-Level Deceptive Response (ODR), an attack that produces
responses accepted by output based verifiers while preserving the original
representation state. By separating output conformity from internal
unlearning, ODR motivates a rethinking of machine unlearning verification,
shifting the audit target from what the model returns to what changes inside
it. We operationalize this shift through Representation Unlearning
Verification (RUV), a gray box verifier that compares intermediate forward
representations in the original and unlearned models. RUV traces a four-stage
response through activation reorganization, representation displacement,
neighborhood disruption, and loss of manifold support. Matched retained
controls distinguish target specific responses from global model drift, while
joint permutation calibration combines the four evidence views into an
adaptive decision. Experiments spanning exact and approximate unlearning,
class and sample level requests, multiple datasets, and diverse architectures
show mean ODR deception success rates of $97.0\%$ and $97.7\%$, respectively.
RUV achieves mean verification accuracies of $98.9\%$ and $94.9\%$, while
rejecting $99.5\%$ and $99.8\%$ of the corresponding ODR submissions.
\end{abstract}

\section{Introduction}
\label{sec_intro}
Removing a training record from storage does not remove its influence from a
learned model. Machine unlearning addresses this problem by updating the model
to eliminate the influence of specified data, preferably without complete
retraining \citep{cao2015towards,ginart2019making,bourtoule2021machine}.
Yet the party requesting unlearning rarely observes the provider's training
process or resulting internal state. An effective unlearning algorithm is
therefore only part of the solution: a verifier must also determine whether a
claimed operation actually occurred.

Machine unlearning still lacks a verification mechanism that is both broadly
applicable and accurate. Process-bound methods either insert traceable signals
before training or require the provider to expose detailed artifacts of the
unlearning process
\citep{sommer2022athena,gao2024verifi,chien2024certified,weng2024proof}. Such
requirements are difficult to satisfy in ordinary model services. A recent study instead inspects gradient changes between the original
and unlearned models, but requires white-box model access and backward
computation, imposing substantial access and computation burdens
\citep{zhou2025truvrf}. Consequently, broadly applicable post
hoc verification still relies mainly on output predictions, using accuracy,
membership inference, and related statistics to infer whether internal
unlearning occurred
\citep{shokri2017membership,carlini2022lira,wang2025tape}.

This reliance assumes that an output resembling unlearning implies a
corresponding model change. Yet a prediction is only the final projection of
the model state, returned by the provider under audit. Confidence, membership,
loss, and other posterior statistics all inherit this evidence source. Can
output conformity and internal unlearning therefore be separated?
\noindent We answer this question constructively with \textbf{Output-Level Deceptive
Response (ODR)}. ODR makes no representation-level parameter update and
produces responses accepted by output verifiers. ODR-Head adjusts only the
classifier for class requests, whereas ODR-Gate returns same-class nonmember
posteriors for registered sample requests without changing the model. Their
success demonstrates that multiple output criteria can report an unlearning
signature even when the computation that produced the original knowledge
remains intact.

We therefore rethink not only where unlearning should be observed, but how an
internal observation becomes verification evidence. Moving the audit boundary
upstream from the returned prediction to intermediate forward representations
exposes the computation from which predictions are formed, but a raw
representation difference is not yet a verifier: ordinary model updates can
also move hidden states, and heterogeneous unlearning objectives need not
produce one universal signature at one layer. Verification consequently
requires three properties. The response must be attributable to the requested
forgotten set rather than model-wide drift, cover complementary stages of
representation formation rather than depend on one statistic, and be calibrated
to the model, data, and request under audit. Our \textbf{Representation
Unlearning Verification (RUV)} framework realizes these properties along the
process by which a representation is formed: activated units define a
computational route; that route produces a feature state; relations among
states establish a local neighborhood; and the neighborhood determines the
sample's support within a learned manifold. RUV follows this progression
through Activation Route Shift (ARS), Representation Shift (RS), Neighborhood
Persistence Gap (NPG), and Manifold Rank Drop (MRD), which trace activation
reorganization, state displacement, neighborhood disruption, and loss of
manifold support. Matched retained controls attribute the response to the
forgotten set, while joint permutation calibration combines corroborating
views without imposing one universal response pattern. Moving upstream also
changes the adversary's burden: output deception can substitute a response at
one exposed interface, whereas passing RUV while preserving utility requires
coordinated, target-attributed changes across multiple stages of the submitted
model's internal computation without being given a target unlearned
representation or audit boundary, turning direct response substitution into a
constrained model-level optimization problem. Appendix~\ref{app:ruv} analyzes
the resulting modification and cost barrier.
Our contributions are:
\begin{itemize}
    \setlength{\itemsep}{0.1em}
    \setlength{\parsep}{0pt}
    \setlength{\topsep}{0.35em}
    \item We introduce ODR-Head and ODR-Gate, constructive attacks that expose
    the vulnerability of output verification for class and sample requests.
    \item We introduce RUV, which relocates the audit boundary to forward
    representations and integrates four stages of the representation response
    chain with retained controls and an adaptive joint decision.
    \item We establish a unified evaluation across exact and approximate
    unlearning, class and sample requests, multiple datasets, and diverse
    architectures, measuring verification, deception, and the role of each
    audit component.
\end{itemize}

\section{Related Work}
\label{sec_related_work}

Some unlearning verifiers establish evidence before or during learning. Active methods
insert triggers, watermarks, canaries, model seeds, or diagnostic sample
relations
\citep{sommer2022athena,guo2024verifying,hu2022membership,gao2024verifi,
wang2026sms,xu2026really,zhang2025duplexguard}. Certified unlearning, replay,
and proof systems bind claims to prescribed optimization, retained records, or
cryptographic artifacts
\citep{chien2024certified,chien2024langevin,nguyen2022mcmc,
thudi2022necessity,weng2024proof,eisenhofer2025verifiable}. These approaches
connect evidence directly to learning, but rely on prearranged signals,
compatible algorithms, retained records, or supporting infrastructure, tying
deployment to particular training and unlearning procedures.

The most broadly deployable verifiers instead infer unlearning from
behavior exposed by prediction interfaces. They test forgotten-set accuracy,
membership, confidence or loss, posterior differences, and task-specific
responses
\citep{shokri2017membership,lu2022fp2mia,carlini2022lira,wang2025tape,
wang2025lifecycle,jeon2026information,vidal2025explainable}. Although these
statistics capture different signatures, they share one evidence boundary:
the final response returned by the audited provider. They consequently share
the premise that an output change consistent with unlearning indicates that
unlearning occurred. ODR targets this premise and shows that this apparent
correspondence can be forged.

Internal evidence offers a different basis for verification. TruVRF measures
the gradient sensitivity of forgotten data
\citep{zhou2025truvrf}, but requires white-box parameters and backward
computation, and remains coupled to the output residual. Outside machine
unlearning verification, representation analyses interpret hidden units,
compare learned spaces, and characterize representation geometry
\citep{bau2017network,kornblith2019similarity,bengio2013representation}. Their
ability to reveal internal computation motivates RUV to move the audit from
returned predictions to forward representations. Yet a representation
difference alone is not an unlearning decision: ordinary updates can also move
hidden states, while heterogeneous unlearning methods need not leave one
universal signature. RUV therefore converts representation observations into
verification evidence through matched retained controls for target attribution,
complementary views of the representation formation chain, and joint
request-specific calibration. It thereby bridges provider-controlled output
transcripts and parameter-level backward inspection without requiring training
logs or a particular unlearning algorithm. Appendix~\ref{app:related} provides
a fuller comparison.

\section{Problem Formulation}
\label{sec_problem}
Let $\mathcal M_o$ be the original model trained on $D$. An unlearning request
specifies the forgotten set $D_u\subset D$, leaving
$D_r=D\setminus D_u$, and the provider submits an unlearned model
$\mathcal M_u$. Unlearning targets are primarily divided into class and sample
requests. A class request removes every sample of specified classes or
identities, whereas a sample request removes designated instances.
\noindent The verifier receives $(\mathcal M_o,\mathcal M_u)$, $D_u$, a matched retained
control $D_r'\subset D_r$, and a reference set $D_{\mathrm{ref}}$ for geometric
audits. It runs both models in an instrumented runtime that exposes intermediate
forward representations, but not parameters, gradients, or training logs.
Access granted to comparison verifiers follows their original protocols. We
assume the runtime faithfully extracts representations from the submitted
model. The verifier retains the audit configuration and intermediate evidence;
deployment attestation and fabrication of the audit channel are outside this
interface.

ODR asks whether a provider can enter the acceptance region of output verifiers
while preserving the original representation state, thereby deceiving existing
output verification. RUV instead verifies an unlearning request by testing
whether the submitted model exhibits corroborated, target-attributed
representation responses on $D_u$ beyond the drift measured on matched retained
data.

\section{Output-Level Deceptive Response}
\label{sec_odr}
ODR constructs accepted outputs without a representation-level parameter
update while preserving retained utility. It exploits the separation between
the computation performed inside a model and the prediction exposed to a
verifier. Class requests alter a shared decision region, so ODR-Head modifies
only the classifier. Sample requests target instances that share their class
structure with retained data, so ODR-Gate intercepts only registered queries.

\paragraph{ODR-Head: class requests.}
Write $\theta_o=(\theta_{\mathrm{rep}}^o,\theta_{\mathrm{head}}^o)$ and freeze
all backbone parameters and inference state. For $K$ classes, let $u_K$ be the
uniform posterior and $\eta(x)$ a perturbed, renormalized posterior of the
\emph{fixed original model}. The forged target and head objective are
\begin{align}
p_{\mathrm{forge}}(x)&=(1-\alpha_{\mathrm{ODR}})u_K+
\alpha_{\mathrm{ODR}}\eta(x),\label{eq:main_forge}\\
\mathcal L_{\mathrm{ODR}}&=
\mathbb E_{x\in D_u}\mathrm{KL}\!\left(p_{\mathrm{forge}}(x)\Vert
p(y\mid x;\theta_{\mathrm{rep}}^o,\theta_{\mathrm{head}})\right)
+\lambda_{\mathrm{ODR}}\mathbb E_{(x,y)\in D_r}\mathrm{CE}(p(y\mid x;\theta),y).
\label{eq:main_odr_loss}
\end{align}
Only the head is optimized. The first term reshapes target predictions; the
second preserves retained utility. Thus, class confusion can be induced
without removing the backbone representations.

\paragraph{ODR-Gate: sample requests.}
Because samples from the same class share a classifier, ODR-Gate instead
intercepts exact registered target queries. Let
$G(x)=\mathbb I[\operatorname{id}(x)\in\mathcal R_u]$, where
$\mathcal R_u=\{\operatorname{id}(x):x\in D_u\}$. A held-out nonmember set
supplies a posterior bank $\mathcal B_c$ for each class $c$. Targets receive
same-class posteriors through a seed-controlled permutation $\pi$, cycling
through the bank if necessary:
\begin{equation}
p_{\mathrm{gate}}(x_i)=G(x_i)q_{\pi(i)}+(1-G(x_i))p_o(x_i),
\qquad q_{\pi(i)}\in\mathcal B_{y_i}.
\label{eq:main_gate}
\end{equation}
This replay reproduces nonmember-like correctness and confidence without a
retraining oracle or parameter updates. Neither construction updates
representation-level parameters, so output acceptance does not establish
internal unlearning. ODR therefore exposes a limitation shared by output
statistics rather than a weakness of one metric: changing the statistic does
not change the provider-controlled interface from which its evidence is drawn.
RUV responds by moving the audit boundary to the forward computation that
produces the output. Appendix~\ref{app:odr} gives the full construction and
diagram.

\section{Representation Unlearning Verification}
\label{sec:ruv_main}
RUV treats unlearning as a response that can propagate through the formation of
an internal representation. Activated units first define the computational
route taken by an input. That route produces a feature state; relations among
states establish a local neighborhood; and the neighborhood determines the
sample's support within a learned manifold. This progression connects
representation learning with the manifold premise that local neighborhoods
encode high dimensional geometry
\citep{bengio2013representation,belkin2001laplacian}. It also explains why no
single feature distance is sufficient: an update can reorganize activations
without a large state displacement, move a state while preserving many
neighbors, or weaken local and manifold support despite a modest absolute
shift.

\begin{figure}[htbp]
\centering
\includegraphics[width=\linewidth]{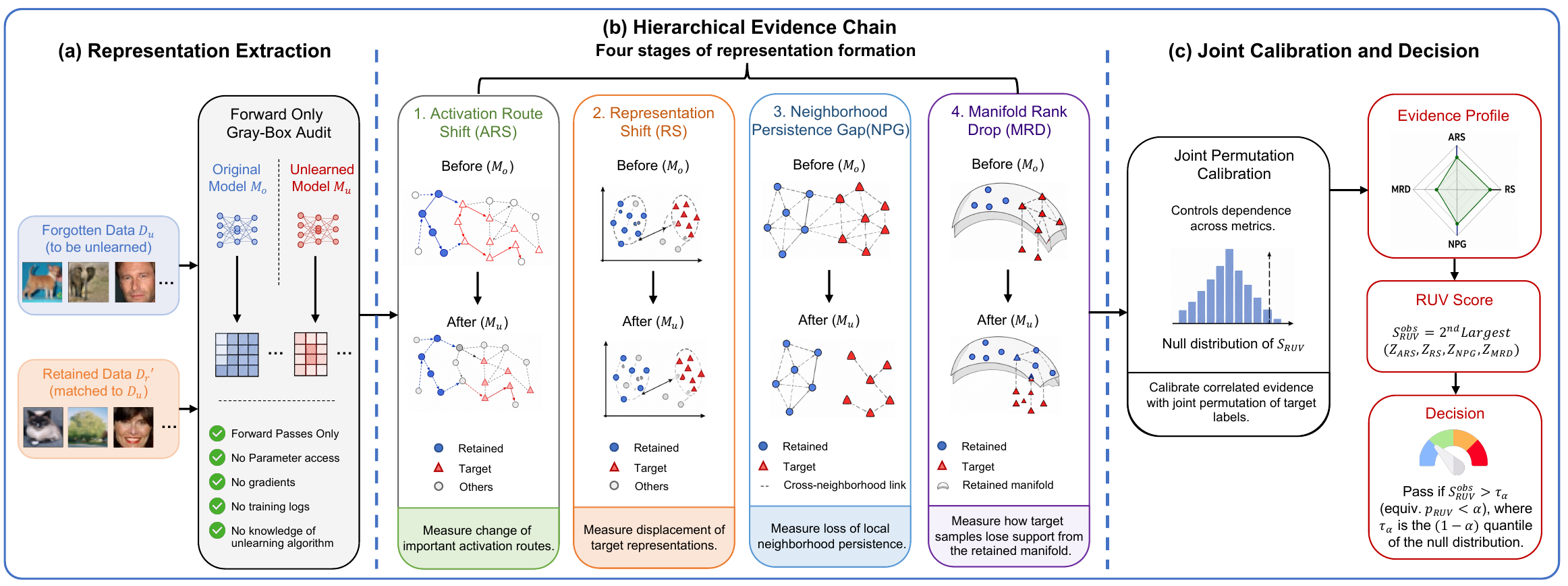}
\caption{RUV follows the representation response from activation routes to
manifold support, then calibrates target-specific evidence against retained
controls.}
\label{fig:ruv_framework}
\end{figure}

RUV follows this chain through four evidence views (Figure~\ref{fig:ruv_framework}):
ARS measures activation reorganization, RS measures state displacement, NPG
measures neighborhood disruption, and MRD measures loss of manifold support.
Different unlearning algorithms may emphasize different stages because they
modify different parameters under different objectives. RUV therefore seeks
corroboration across the chain rather than requiring one universal signature.
Every score is computed per sample before group averaging, and matched retained
samples separate a response concentrated on $D_u$ from model-wide drift.
This design raises the manipulation target from a returned posterior to model
computation. Unlike output deception, which can directly substitute a response
at one exposed interface, passing RUV requires target-specific, cross-view
model updates while retained utility remains stable. The verifier does not
expose its controls, intermediate evidence, or request-specific boundary before
submission. Constructing a reliable internal target can therefore require a
reference unlearning execution, followed by additional optimization to preserve
the information being concealed. Appendix~\ref{app:ruv} analyzes this
constraint and cost asymmetry, while Appendix~\ref{app:theory} formalizes why
output evidence can be matched and why RUV detects a corroborated target
response.

\vspace{-10pt}

\paragraph{Representation access and selection.}
For block $l$, a fixed operator $\mathcal P_l$ maps raw output $h^{(l)}$ to the
audit vector
\begin{equation}
z^{(l)}(x;\mathcal M)=
\operatorname{vec}\!\left(\mathcal P_l[h^{(l)}(x;\mathcal M)]\right).
\label{eq:main_representation}
\end{equation}
The same deterministic operator is applied to $\mathcal M_o$ and
$\mathcal M_u$. CNN blocks use spatial pooling, Transformer blocks use their
summary token, and the preclassifier block $l_{\mathrm{pre}}$ uses the native
classifier input. With ordered representation blocks $\mathcal L$, RUV fixes
the audited sets by request granularity:
\begin{equation}
\mathcal L_{\mathrm{class}}=\{l_{\mathrm{pre}}\},
\qquad \mathcal L_{\mathrm{sample}}=\mathcal L\setminus\{l_{\mathrm{pre}}\}.
\label{eq:main_layers}
\end{equation}
The first targets terminal class structure
\citep{yosinski2014transferable,papyan2020prevalence}; the latter captures
instance traces distributed across nonterminal blocks
\citep{maini2023localized}. Appendix~\ref{sec:ruv_chain} specifies the pooling
operator, block registry, and architecture-specific extraction rules.

\subsection{Four Complementary Evidence Views}
Below, superscripts $o,u$ indicate $\mathcal M_o,\mathcal M_u$ at the same
block and input. Vector geometry uses cosine distance $d_{\cos}$; comparisons
of index sets use Jaccard similarity $J(A,B)=|A\cap B|/|A\cup B|$, with
$J(\varnothing,\varnothing)=1$. These metrics address different objects and
are not interchangeable.

\paragraph{Activation Route Shift (ARS).}
Let $A^m$ index the top $q_{\mathrm{top}}$ fraction of magnitudes in $z^m$,
and $B^m=\{i:z_i^m>\gamma_{\mathrm{act}}\}$. ARS measures changes in both
dominant and thresholded activation coordinates:
\begin{equation}
s_{\mathrm{ARS}}^{(l)}(x)=\omega_b[1-J(B^o,B^u)]
+(1-\omega_b)[1-J(A^o,A^u)].\label{eq:main_ars}
\end{equation}
The two sets distinguish activation flips from redistribution among dominant
coordinates, connecting this view to the semantic structure carried by hidden
units \citep{bau2017network}.

\paragraph{Representation Shift (RS).}
RS measures displacement of the resulting feature state:
\begin{equation}
s_{\mathrm{RS}}^{(l)}(x)=d_{\cos}(z^o,z^u)
=1-\frac{(z^o)^\top z^u}
{\max(\|z^o\|_2,\epsilon_d)\max(\|z^u\|_2,\epsilon_d)}.
\label{eq:main_rs}
\end{equation}
Unlike ARS, it considers the full vector rather than activation membership.
Cosine distance emphasizes direction and reduces sensitivity to blockwise
activation scale, but displacement alone does not describe relational change.

\paragraph{Neighborhood Persistence Gap (NPG).}
Let $\mathcal N^m_k(x)$ contain the identities of the $k_{\mathrm{NPG}}$
nearest reference samples under $d_{\cos}$ in model $m$. NPG measures loss of
neighbor persistence:
\begin{equation}
s_{\mathrm{NPG}}^{(l)}(x)=1-J(\mathcal N^o_k(x),\mathcal N^u_k(x)).
\label{eq:main_npg}
\end{equation}
Reference identities stay fixed across models, although their representations
are recomputed. A sample can move little yet change its neighbors; conversely,
a common geometry preserving transformation can move coordinates without
disrupting neighborhood identities. NPG therefore complements RS rather than
averaging the same distances again.

\paragraph{Manifold Rank Drop (MRD).}
Let $\rho_k^{(l)}(x;\mathcal M)$ be mean cosine affinity to the
$k_{\mathrm{MRD}}$ nearest label compatible reference samples. Calibrate it
against the reference affinity distribution $\mathcal C_k^{(l)}(\mathcal M)$,
with self matches excluded:
\begin{align}
r_k^{(l)}(x;\mathcal M)&=\frac{1}{|\mathcal C_k^{(l)}(\mathcal M)|}
\sum_{c\in\mathcal C_k^{(l)}(\mathcal M)}\mathbb I[c\leq\rho_k^{(l)}(x;\mathcal M)],\\
s_{\mathrm{MRD}}^{(l)}(x)&=r_k^{(l)}(x;\mathcal M_o)-r_k^{(l)}(x;\mathcal M_u).
\label{eq:main_mrd}
\end{align}
MRD asks whether reference support declines, even when new neighbors replace
old ones. Sample audits use retained references; class audits require class
compatible held-out references because the target class is absent from $D_r$.
Reference affinities are recomputed in each model, so the rank measures support
relative to that model's geometry rather than against a fixed original scale.
Replacing neighbors with equally supportive alternatives can therefore increase
NPG without reducing MRD, whereas relative support can decline even when some
original neighbors remain. Appendix~\ref{app:ruv} gives the full affinity and
rank construction, which adapts neighborhood support as a representation
reliability signal \citep{jiang2018trust}.

\subsection{Target Attribution and Joint Decision}
Let $j\in\{\mathrm{ARS},\mathrm{RS},\mathrm{NPG},\mathrm{MRD}\}$.
For $\mathcal X=D_u\cup D_r'$, standardize each block's scores using its pooled
mean $\mu_{j,l}^{\mathrm{pool}}$ and population standard deviation
$\sigma_{j,l}^{\mathrm{pool}}$, then equally average audited blocks:
\begin{equation}
S_j(x)=\frac{1}{|\mathcal L_g|}\sum_{l\in\mathcal L_g}
\frac{s_j^{(l)}(x)-\mu_{j,l}^{\mathrm{pool}}}
{\sigma_{j,l}^{\mathrm{pool}}+\epsilon_s},
\qquad G_j^{\mathrm{obs}}=\operatorname{mean}_{D_u}S_j-
\operatorname{mean}_{D_r'}S_j.
\label{eq:main_gap}
\end{equation}
Pooling is group label free. Retained controls are sampled without replacement,
with $|D_r'|=|D_u|$, and matched by class for sample requests.
The gap attributes change to the request rather
than to drift affecting the whole model. Positive gaps alone are insufficient
because random variation differs between views and audits.

RUV therefore jointly permutes the audit group labels of $\mathcal X$ while
keeping task labels, model outputs, and each sample's four-view vector fixed.
The same permutation is used for all views, preserving their empirical
dependence \citep{romano2005exact}. Sample level permutations are stratified by
class; class level audits use size preserving pseudo target assignments over
pooled audit samples. Paired representations and any augmented views remain
in the same sample block. From $B$ permuted gaps $G_j^{(b)}$, compute the null
mean $\mu_j^{(0)}$ and sample standard deviation $\sigma_j^{(0)}$, and define
\begin{equation}
Z_j^{(t)}=
\begin{cases}
(G_j^{(t)}-\mu_j^{(0)})/\sigma_j^{(0)},&\sigma_j^{(0)}>\epsilon_s,\\
0,&\text{otherwise},
\end{cases}
\quad t\in\{\mathrm{obs},1,\ldots,B\}.
\label{eq:main_z}
\end{equation}
Each view is now expressed in units of its own null variation. The second
largest effect requires corroboration from at least two views without learned
weights or requiring all four to respond equally:
\begin{equation}
S_{\mathrm{RUV}}^{(t)}=\operatorname{2ndLargest}_j Z_j^{(t)},\quad
\tau_\alpha=Q_{1-\alpha}\{S_{\mathrm{RUV}}^{(b)}\}_{b=1}^{B},\quad
E_{\mathrm{RUV}}=\mathbb I[\Delta_{\mathrm{RUV}}>0],
\label{eq:main_decision}
\end{equation}
where $\Delta_{\mathrm{RUV}}=S_{\mathrm{RUV}}^{\mathrm{obs}}-\tau_\alpha$.
The quantile uses order rank $\min\{B,\lceil(B+1)(1-\alpha)\rceil\}$.
Every trial is judged against its own boundary before results are aggregated;
an average boundary never replaces per-trial decisions.
Appendix~\ref{app:ruv} gives the complete definitions.

The four views are therefore neither four independent acceptance tests nor
four copies of a feature distance. ARS and RS inspect computation and state;
NPG and MRD inspect neighborhood identity and relative support. Retained
comparison supplies target attribution, while joint calibration supplies the
request-specific fluctuation boundary. This division of roles allows different
unlearning mechanisms to yield different evidence profiles under the same
decision rule. Appendix~\ref{app:theory} establishes the corresponding
properties of output-only indistinguishability, retained-control attribution,
multi-view corroboration, and permutation-calibrated detection.

\section{Experiments}
\label{sec:experiments}
\label{sec_experiments}
\subsection{Experimental Settings}
\textbf{Datasets and models.}
We pair CIFAR10, CIFAR100, and FaceScrub with ResNet18, ResNet34, and ResNet50,
respectively \citep{krizhevsky2009learning,ng2014facescrub,he2016resnet}.
CIFAR10 and CIFAR100 cover increasing category complexity, while FaceScrub
extends the evaluation to identity recognition. Additional architectures are
evaluated in Section~\ref{sec:architecture_robustness}.

\textbf{Unlearning settings.}
Main experiments remove one class or identity for class requests and use
stratified sampling with $\rho=10\%$ for sample requests; other request scales
are studied in Appendix~\ref{sec:request_scale_robustness}. Retraining and SISA
provide exact baselines, while Amnesiac, SCRUB, SalUn, and SSD cover distinct
approximate mechanisms \citep{bourtoule2021machine,graves2021amnesiac,
kurmanji2023scrub,fan2024salun,foster2024ssd}. Each setting contains 100
submissions from ten original models and ten fixed-manifest requests per model.

\textbf{Verification baselines.}
We adopt output-based verifiers that cover complementary signals: Accuracy
uses predictive accuracy, LiRA performs membership inference, and TAPE measures
posterior differences. We also compare with TruVRF, which audits internal model
changes through gradients \citep{carlini2022lira,wang2025tape,zhou2025truvrf}.

\textbf{Evaluation metrics.}
Verification accuracy (VA) is the fraction of configured unlearning executions
accepted and adversarial submissions rejected under this controlled protocol.
ODR deception success rate (DSR) is the fraction of ODR submissions accepted by
an output verifier; detection rate is its complement.

Implementation details, repeated-audit aggregation, and baseline calibration
are provided in Appendix~\ref{app:setup}.

\subsection{Main Results}
\paragraph{Output-level deception.}
\begin{table}[!ht]
\centering
\caption{ODR deception success rate (C: class; S: sample). Each point estimate
is the success count over 1000 binary audits; uncertainty is the standard error
over ten original-model summaries.}
\label{tab:odr_dsr}
\fontsize{7.8}{9.2}\selectfont
\setlength{\tabcolsep}{3.0pt}
\renewcommand{\arraystretch}{1.08}
\begin{tabular*}{\linewidth}{@{\extracolsep{\fill}}lcccccc@{}}
\toprule
& \multicolumn{2}{c}{CIFAR10}
& \multicolumn{2}{c}{CIFAR100}
& \multicolumn{2}{c}{FaceScrub} \\
\cmidrule(lr){2-3}\cmidrule(lr){4-5}\cmidrule(lr){6-7}
Verifier & Class & Sample & Class & Sample & Class & Sample \\
\midrule
Accuracy
& \textbf{0.992 $\pm$ 0.003} & \textbf{0.991 $\pm$ 0.003}
& \textbf{0.988 $\pm$ 0.005} & \textbf{0.987 $\pm$ 0.005}
& \textbf{0.996 $\pm$ 0.003} & \textbf{0.993 $\pm$ 0.003} \\
LiRA
& 0.958 $\pm$ 0.015 & 0.969 $\pm$ 0.008
& 0.948 $\pm$ 0.012 & 0.958 $\pm$ 0.011
& 0.942 $\pm$ 0.011 & 0.965 $\pm$ 0.009 \\
TAPE
& 0.972 $\pm$ 0.012 & 0.978 $\pm$ 0.007
& 0.964 $\pm$ 0.008 & 0.972 $\pm$ 0.009
& 0.968 $\pm$ 0.008 & 0.980 $\pm$ 0.008 \\
\bottomrule
\end{tabular*}
\end{table}

We first ask whether a provider can satisfy several output-based verifiers
without performing a representation-level parameter update. We submit ODR-Head
for class requests and ODR-Gate for sample requests to Accuracy, LiRA, and TAPE
on all three datasets. Table~\ref{tab:odr_dsr} shows mean DSRs of $0.970$ and
$0.977$, respectively, with even the lowest rates remaining $0.942$ and
$0.958$. The result spans accuracy, membership, and posterior-difference
criteria while preserving retained accuracy: ODR-Head suppresses class
accuracy, whereas ODR-Gate replays the high-confidence same-class nonmember
behavior produced by sample-level Retraining. These verifiers inspect different
statistics, but all derive them from the same provider-controlled response;
ODR therefore produces output conformity while leaving the computation that
formed the original representation unmodified. Appendix~\ref{app:odr_results}
reports complete prediction behavior, retained utility, and construction cost.

\begin{table}[!ht]
\centering
\caption{Verification accuracy across methods and request levels. Each point
estimate is the success count over 1000 binary audits; uncertainty is the
standard error over ten original-model summaries.}
\label{tab:verification_main}\label{tab:class_verification}\label{tab:sample_verification}
\fontsize{7.5}{9}\selectfont
\setlength{\tabcolsep}{1.8pt}
\renewcommand{\arraystretch}{1.12}
\textbf{(a) Class level}\par\smallskip
\begin{tabular*}{\linewidth}{@{\extracolsep{\fill}}ccccccccc@{}}
\toprule
\multicolumn{1}{c}{Dataset} & \multicolumn{1}{c}{Verifier} & Retraining & SISA & Amnesiac & SCRUB & SalUn & SSD & \textbf{ODR} \\
\midrule
\multirow{5}{*}{CIFAR10} & Accuracy & 0.964 $\pm$ 0.010 & 0.950 $\pm$ 0.013 & 0.888 $\pm$ 0.015 & 0.928 $\pm$ 0.024 & 0.906 $\pm$ 0.035 & 0.860 $\pm$ 0.031 & \textbf{0.008 $\pm$ 0.003} \\
 & LiRA & 0.954 $\pm$ 0.014 & 0.932 $\pm$ 0.013 & 0.902 $\pm$ 0.023 & 0.910 $\pm$ 0.023 & 0.918 $\pm$ 0.016 & 0.874 $\pm$ 0.017 & \textbf{0.042 $\pm$ 0.015} \\
 & TAPE & 0.976 $\pm$ 0.009 & 0.962 $\pm$ 0.012 & 0.928 $\pm$ 0.014 & 0.950 $\pm$ 0.011 & 0.936 $\pm$ 0.029 & 0.904 $\pm$ 0.018 & \textbf{0.028 $\pm$ 0.012} \\
 & TruVRF & 0.948 $\pm$ 0.008 & 0.922 $\pm$ 0.027 & 0.866 $\pm$ 0.027 & 0.900 $\pm$ 0.020 & 0.884 $\pm$ 0.017 & 0.840 $\pm$ 0.022 & \textbf{0.138 $\pm$ 0.034} \\
 & \textbf{RUV} & \textbf{0.996 $\pm$ 0.003} & \textbf{0.994 $\pm$ 0.003} & \textbf{0.986 $\pm$ 0.005} & \textbf{0.992 $\pm$ 0.004} & \textbf{0.990 $\pm$ 0.005} & \textbf{0.984 $\pm$ 0.006} & \textbf{0.998 $\pm$ 0.002} \\
\midrule
\multirow{5}{*}{CIFAR100} & Accuracy & 0.938 $\pm$ 0.026 & 0.920 $\pm$ 0.018 & 0.852 $\pm$ 0.029 & 0.898 $\pm$ 0.031 & 0.880 $\pm$ 0.022 & 0.832 $\pm$ 0.026 & \textbf{0.012 $\pm$ 0.005} \\
 & LiRA & 0.928 $\pm$ 0.016 & 0.910 $\pm$ 0.027 & 0.868 $\pm$ 0.027 & 0.878 $\pm$ 0.029 & 0.894 $\pm$ 0.012 & 0.836 $\pm$ 0.017 & \textbf{0.052 $\pm$ 0.012} \\
 & TAPE & 0.958 $\pm$ 0.010 & 0.930 $\pm$ 0.019 & 0.894 $\pm$ 0.023 & 0.914 $\pm$ 0.009 & 0.910 $\pm$ 0.018 & 0.880 $\pm$ 0.032 & \textbf{0.036 $\pm$ 0.008} \\
 & TruVRF & 0.924 $\pm$ 0.018 & 0.896 $\pm$ 0.020 & 0.828 $\pm$ 0.040 & 0.866 $\pm$ 0.037 & 0.846 $\pm$ 0.030 & 0.802 $\pm$ 0.032 & \textbf{0.174 $\pm$ 0.035} \\
 & \textbf{RUV} & \textbf{0.996 $\pm$ 0.003} & \textbf{0.992 $\pm$ 0.004} & \textbf{0.982 $\pm$ 0.010} & \textbf{0.990 $\pm$ 0.004} & \textbf{0.988 $\pm$ 0.005} & \textbf{0.980 $\pm$ 0.008} & \textbf{0.996 $\pm$ 0.004} \\
\midrule
\multirow{5}{*}{FaceScrub} & Accuracy & 0.952 $\pm$ 0.016 & 0.934 $\pm$ 0.013 & 0.882 $\pm$ 0.032 & 0.916 $\pm$ 0.015 & 0.886 $\pm$ 0.033 & 0.854 $\pm$ 0.027 & \textbf{0.004 $\pm$ 0.003} \\
 & LiRA & 0.944 $\pm$ 0.023 & 0.934 $\pm$ 0.022 & 0.888 $\pm$ 0.025 & 0.902 $\pm$ 0.021 & 0.906 $\pm$ 0.020 & 0.858 $\pm$ 0.024 & \textbf{0.058 $\pm$ 0.011} \\
 & TAPE & 0.962 $\pm$ 0.011 & 0.960 $\pm$ 0.012 & 0.918 $\pm$ 0.022 & 0.936 $\pm$ 0.028 & 0.938 $\pm$ 0.011 & 0.890 $\pm$ 0.014 & \textbf{0.032 $\pm$ 0.008} \\
 & TruVRF & 0.946 $\pm$ 0.023 & 0.908 $\pm$ 0.018 & 0.848 $\pm$ 0.023 & 0.882 $\pm$ 0.029 & 0.874 $\pm$ 0.021 & 0.816 $\pm$ 0.024 & \textbf{0.158 $\pm$ 0.030} \\
 & \textbf{RUV} & \textbf{0.992 $\pm$ 0.006} & \textbf{0.994 $\pm$ 0.006} & \textbf{0.984 $\pm$ 0.007} & \textbf{0.990 $\pm$ 0.006} & \textbf{0.986 $\pm$ 0.006} & \textbf{0.982 $\pm$ 0.004} & \textbf{0.990 $\pm$ 0.004} \\
\bottomrule
\end{tabular*}
\par\medskip
\textbf{(b) Sample level}\par\smallskip
\begin{tabular*}{\linewidth}{@{\extracolsep{\fill}}ccccccccc@{}}
\toprule
\multicolumn{1}{c}{Dataset} & \multicolumn{1}{c}{Verifier} & Retraining & SISA & Amnesiac & SCRUB & SalUn & SSD & \textbf{ODR} \\
\midrule
\multirow{5}{*}{CIFAR10} & Accuracy & 0.931 $\pm$ 0.015 & 0.917 $\pm$ 0.024 & 0.849 $\pm$ 0.040 & 0.893 $\pm$ 0.028 & 0.871 $\pm$ 0.020 & 0.823 $\pm$ 0.016 & \textbf{0.009 $\pm$ 0.003} \\
 & LiRA & 0.924 $\pm$ 0.020 & 0.901 $\pm$ 0.022 & 0.865 $\pm$ 0.018 & 0.879 $\pm$ 0.025 & 0.884 $\pm$ 0.017 & 0.833 $\pm$ 0.027 & \textbf{0.031 $\pm$ 0.008} \\
 & TAPE & 0.952 $\pm$ 0.013 & 0.938 $\pm$ 0.013 & 0.899 $\pm$ 0.032 & 0.919 $\pm$ 0.026 & 0.909 $\pm$ 0.016 & 0.868 $\pm$ 0.015 & \textbf{0.022 $\pm$ 0.007} \\
 & TruVRF & 0.918 $\pm$ 0.017 & 0.884 $\pm$ 0.036 & 0.821 $\pm$ 0.027 & 0.857 $\pm$ 0.032 & 0.847 $\pm$ 0.042 & 0.794 $\pm$ 0.033 & \textbf{0.784 $\pm$ 0.029} \\
 & \textbf{RUV} & \textbf{0.961 $\pm$ 0.003} & \textbf{0.957 $\pm$ 0.006} & \textbf{0.943 $\pm$ 0.007} & \textbf{0.954 $\pm$ 0.003} & \textbf{0.950 $\pm$ 0.007} & \textbf{0.938 $\pm$ 0.008} & \textbf{0.999 $\pm$ 0.001} \\
\midrule
\multirow{5}{*}{CIFAR100} & Accuracy & 0.904 $\pm$ 0.031 & 0.881 $\pm$ 0.020 & 0.808 $\pm$ 0.038 & 0.856 $\pm$ 0.026 & 0.836 $\pm$ 0.026 & 0.785 $\pm$ 0.038 & \textbf{0.013 $\pm$ 0.005} \\
 & LiRA & 0.898 $\pm$ 0.015 & 0.872 $\pm$ 0.029 & 0.824 $\pm$ 0.023 & 0.842 $\pm$ 0.023 & 0.853 $\pm$ 0.029 & 0.790 $\pm$ 0.033 & \textbf{0.042 $\pm$ 0.011} \\
 & TAPE & 0.925 $\pm$ 0.022 & 0.894 $\pm$ 0.014 & 0.855 $\pm$ 0.031 & 0.877 $\pm$ 0.019 & 0.869 $\pm$ 0.027 & 0.838 $\pm$ 0.036 & \textbf{0.028 $\pm$ 0.009} \\
 & TruVRF & 0.882 $\pm$ 0.019 & 0.855 $\pm$ 0.024 & 0.781 $\pm$ 0.046 & 0.822 $\pm$ 0.019 & 0.799 $\pm$ 0.033 & 0.755 $\pm$ 0.042 & \textbf{0.778 $\pm$ 0.035} \\
 & \textbf{RUV} & \textbf{0.957 $\pm$ 0.005} & \textbf{0.952 $\pm$ 0.009} & \textbf{0.936 $\pm$ 0.005} & \textbf{0.949 $\pm$ 0.005} & \textbf{0.946 $\pm$ 0.006} & \textbf{0.932 $\pm$ 0.006} & \textbf{0.998 $\pm$ 0.002} \\
\midrule
\multirow{5}{*}{FaceScrub} & Accuracy & 0.920 $\pm$ 0.014 & 0.903 $\pm$ 0.019 & 0.841 $\pm$ 0.042 & 0.881 $\pm$ 0.033 & 0.849 $\pm$ 0.035 & 0.813 $\pm$ 0.028 & \textbf{0.007 $\pm$ 0.003} \\
 & LiRA & 0.913 $\pm$ 0.019 & 0.899 $\pm$ 0.029 & 0.854 $\pm$ 0.017 & 0.868 $\pm$ 0.033 & 0.867 $\pm$ 0.023 & 0.818 $\pm$ 0.033 & \textbf{0.035 $\pm$ 0.009} \\
 & TAPE & 0.936 $\pm$ 0.013 & 0.932 $\pm$ 0.021 & 0.884 $\pm$ 0.028 & 0.908 $\pm$ 0.023 & 0.904 $\pm$ 0.026 & 0.853 $\pm$ 0.015 & \textbf{0.020 $\pm$ 0.008} \\
 & TruVRF & 0.910 $\pm$ 0.013 & 0.865 $\pm$ 0.021 & 0.804 $\pm$ 0.039 & 0.840 $\pm$ 0.020 & 0.831 $\pm$ 0.016 & 0.772 $\pm$ 0.035 & \textbf{0.826 $\pm$ 0.031} \\
 & \textbf{RUV} & \textbf{0.963 $\pm$ 0.003} & \textbf{0.959 $\pm$ 0.004} & \textbf{0.945 $\pm$ 0.009} & \textbf{0.953 $\pm$ 0.004} & \textbf{0.948 $\pm$ 0.008} & \textbf{0.940 $\pm$ 0.010} & \textbf{0.997 $\pm$ 0.002} \\
\bottomrule
\end{tabular*}
\end{table}

\FloatBarrier
\paragraph{RUV performance.}

We next test whether RUV verifies exact and approximate unlearning at both
request levels while rejecting submissions that only reproduce expected
outputs. Table~\ref{tab:verification_main} compares all five verifiers on the
same model pairs, request manifests, and submissions; unlearning columns report
correct acceptance and the ODR column reports correct rejection. RUV leads
every method column: its pooled class-level accuracies are $0.995$ for
Retraining, $0.993$ for SISA, and $0.982$--$0.991$ for approximate methods,
while the sample-level counterparts are $0.960$, $0.956$, and
$0.937$--$0.952$. Averaged across datasets and methods, RUV reaches $0.989$ and
$0.949$, exceeding TAPE, the strongest comparison verifier, by $5.8$ and $5.1$
percentage points. RUV also rejects ODR-Head and ODR-Gate at $0.995$ and
$0.998$; the output verifiers reject at most $0.051$ and $0.036$, while TruVRF
reaches $0.157$ and $0.796$ because its backward sensitivity is recomputed from
the submitted model. RUV thus remains effective from broad exact-retraining
responses to narrower approximate updates without depending on one unlearning
mechanism. Its lower sample-level values reflect weaker, distributed instance
responses within shared class structure, while its ODR rejection demonstrates
the advantage of auditing the internal computation preceding the output.
Retraining and SISA lead because rebuilding without $D_u$ changes the
representation chain coherently, while narrower approximate updates remain
detectable. Class requests are easier because category removal reorganizes
terminal geometry, whereas sample effects remain distributed across
nonterminal representations.
Appendix~\ref{app:ruv_results} reports the complete margins, evidence profiles,
and overhead: on CIFAR10/ResNet18, RUV takes $4.37$ seconds and $1684$ MB,
compared with $91.2$ seconds and $2528$ MB for TruVRF.

\label{sec:completion_sensitivity}

\subsection{Ablation Studies}

\paragraph{Request completion.}
\begin{figure}[t]
    \centering
    \includegraphics[width=0.92\linewidth]{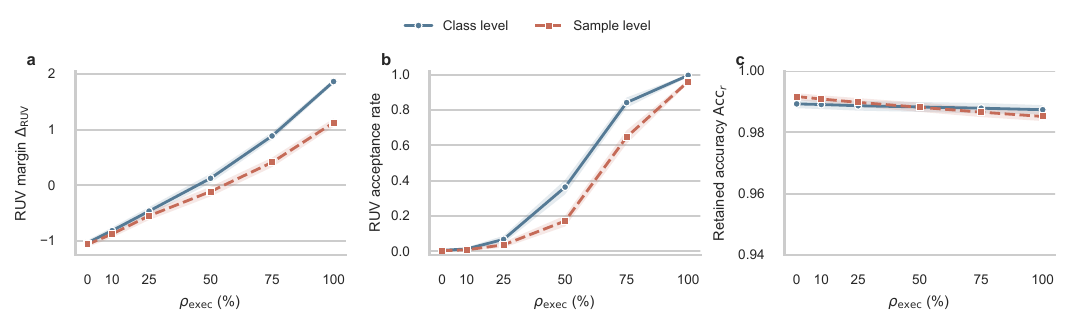}
    \caption{RUV response as the executed fraction of a fixed request increases.}
    \label{fig:completion_continuum}
\end{figure}
To determine whether RUV follows request fulfillment rather than merely
recognizing that Retraining was invoked, we fix the CIFAR10/ResNet18 original
models, complete request $D_u$, training procedure, audit inputs, retained
controls, and references, and vary only the executed subset
$D_u^{(\rho_{\mathrm{exec}})}\subseteq D_u$. RUV always audits the complete
$D_u$, including when the same training procedure excludes no requested sample.
Figure~\ref{fig:completion_continuum} shows an ordered response: from zero to
complete execution, class-level margin and acceptance rise from $-1.032$ and
$0.006$ to $1.863$ and $0.996$, while their sample-level counterparts rise from
$-1.061$ and $0.004$ to $1.125$ and $0.961$. Acceptance remains only
$0.068/0.037$ at $25\%$ execution but reaches $0.842/0.647$ at $75\%$.
Retained accuracy changes by only $0.19/0.65$ percentage points and ends at
$0.987/0.985$. The synchronized progression therefore tracks how much of the
fixed request is carried out rather than routine identity or indiscriminate
utility loss.

\paragraph{Target attribution.}
\label{sec:target_attribution_results}
\begin{wraptable}[9]{r}{0.50\textwidth}
\vspace{-6pt}
\centering
\caption{Target attribution and non-unlearning controls.}
\label{tab:target_attribution_ablation}
\fontsize{7.5}{9}\selectfont
\setlength{\tabcolsep}{3.0pt}
\renewcommand{\arraystretch}{1.02}
\begin{tabular*}{\linewidth}{@{\extracolsep{\fill}}ccc@{}}
\toprule
Setting & Class & Sample \\
\midrule
\textbf{Standard audit} & \textbf{0.991 $\pm$ 0.006} & \textbf{0.957 $\pm$ 0.008} \\
Benign drift control & 0.989 $\pm$ 0.014 & 0.947 $\pm$ 0.017 \\
Wrong target unlearning control & 0.976 $\pm$ 0.035 & 0.937 $\pm$ 0.022 \\
Same target fine-tuning control & 0.968 $\pm$ 0.028 & 0.929 $\pm$ 0.024 \\
Without retained control & 0.842 $\pm$ 0.025 & 0.774 $\pm$ 0.024 \\
\bottomrule
\end{tabular*}
\end{wraptable}

We test whether RUV accepts arbitrary model movement or any change on the
correct $D_u$ by comparing the standard audit with a variant without matched
retained controls and three non-unlearning updates: continued full-data
training, unlearning an equally sized disjoint $D_w\subset D_r$ while auditing
$D_u$, and ten epochs of correct-label fine-tuning on $D_u$. The last control
changes target representations while preserving and strengthening their
information. Table~\ref{tab:target_attribution_ablation} shows that removing
the retained control reduces class/sample VA from $0.991/0.957$ to
$0.842/0.774$. Standard RUV correctly rejects benign drift at $0.989/0.947$,
wrong-target unlearning at $0.976/0.937$, and same-target fine-tuning at
$0.968/0.929$; in the last case, forgotten-set accuracy is $0.998/0.997$,
confidence is $0.992/0.990$, and retained accuracy remains $0.981/0.986$.
Thus, generic drift, an unlearning-shaped update elsewhere, and a target
specific change that retains target knowledge are all insufficient: accepted
evidence must be attributable to the requested set and consistent with an
unlearning response.

\WFclear

\paragraph{Evidence components.}
\begin{wraptable}[10]{r}{0.50\textwidth}
\vspace{-3pt}
\centering
\caption{Ablation of RUV evidence views.}
\label{tab:component_ablation}
\fontsize{7.5}{8}\selectfont
\setlength{\tabcolsep}{3.0pt}
\renewcommand{\arraystretch}{0.96}
\begin{tabular*}{\linewidth}{@{\extracolsep{\fill}}ccc@{}}
\toprule
Variant & Class & Sample \\
\midrule
ARS only & 0.838 $\pm$ 0.022 & 0.714 $\pm$ 0.029 \\
RS only & 0.873 $\pm$ 0.019 & 0.782 $\pm$ 0.025 \\
NPG only & 0.916 $\pm$ 0.016 & 0.848 $\pm$ 0.022 \\
MRD only & 0.894 $\pm$ 0.020 & 0.702 $\pm$ 0.031 \\
without ARS & 0.982 $\pm$ 0.009 & 0.941 $\pm$ 0.013 \\
without RS & 0.975 $\pm$ 0.011 & 0.928 $\pm$ 0.015 \\
without NPG & 0.961 $\pm$ 0.014 & 0.899 $\pm$ 0.019 \\
without MRD & 0.969 $\pm$ 0.012 & 0.932 $\pm$ 0.016 \\
\textbf{Full RUV} & \textbf{0.991 $\pm$ 0.006} & \textbf{0.957 $\pm$ 0.008} \\
\bottomrule
\end{tabular*}
\end{wraptable}

To test whether the four views are interchangeable, we use each view alone and
remove each from the complete verifier on the same 100 class and 100 sample
requests. Table~\ref{tab:component_ablation} identifies NPG as the strongest
individual view, yet it remains $7.5$ and $10.9$ percentage points below full
RUV. Every leave-one-out variant also lowers accuracy; the largest sample-level
loss occurs without NPG, while removing MRD still reduces sample VA from
$0.957$ to $0.932$. Activation organization, coordinate displacement, neighbor
identity, and manifold support therefore capture successive but nonredundant
consequences of an internal update, and no single statistic reproduces their
corroborated evidence.
\WFclear

\paragraph{Evidence aggregation.}
\begin{wraptable}[7]{r}{0.47\textwidth}
\vspace{-6pt}
\centering
\caption{Ablation of evidence aggregation.}
\label{tab:aggregation_ablation}
\fontsize{7.5}{9}\selectfont
\setlength{\tabcolsep}{3.0pt}
\renewcommand{\arraystretch}{1.08}
\begin{tabular*}{\linewidth}{@{\extracolsep{\fill}}ccc@{}}
\toprule
Aggregation & Class & Sample \\
\midrule
Maximum & 0.945 $\pm$ 0.014 & 0.901 $\pm$ 0.017 \\
Mean & 0.972 $\pm$ 0.011 & 0.931 $\pm$ 0.014 \\
Minimum & 0.919 $\pm$ 0.016 & 0.864 $\pm$ 0.020 \\
\textbf{Second largest} & \textbf{0.991 $\pm$ 0.006} & \textbf{0.957 $\pm$ 0.008} \\
\bottomrule
\end{tabular*}
\end{wraptable}

We isolate the decision rule by holding all evidence scores and permutation
samples fixed and replacing only the aggregation operator. Maximum, mean, and
minimum aggregation reach class/sample VA of $0.945/0.901$, $0.972/0.931$, and
$0.919/0.864$, whereas the second-largest effect reaches $0.991/0.957$
(Table~\ref{tab:aggregation_ablation}). A maximum can accept one anomalous view,
a minimum requires every mechanism to affect the full chain, and a mean dilutes
localized evidence. The second-largest rule instead requires two corroborating
views without fixing their identities or learning weights, matching the
heterogeneous propagation patterns observed across unlearning methods.
\WFclear

\paragraph{Architecture robustness.}
\label{sec:architecture_robustness}
\begin{wrapfigure}[12]{r}{0.46\textwidth}
    \vspace{-7pt}
    \centering
    \includegraphics[width=\linewidth]{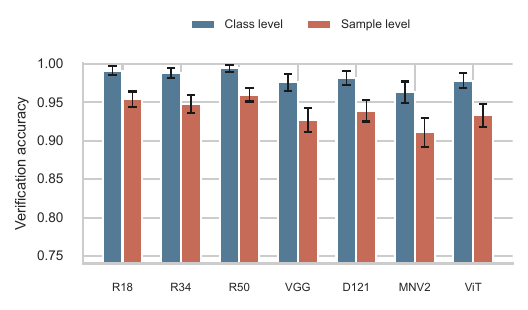}
    \caption{RUV accuracy across model architectures.}
    \label{fig:architecture_robustness}
\end{wrapfigure}
We finally test whether RUV depends on one convolutional design by applying the
same request construction, representation selection, retained controls, and
joint calibration to seven CNN and Transformer backbones on CIFAR10. Class VA
remains within $0.963$--$0.994$ and sample VA within $0.911$--$0.960$;
ViT-S/16 reaches $0.978/0.933$, while MobileNetV2 reaches $0.963/0.911$.
The consistent results across residual, densely connected, lightweight, and
Transformer families show that RUV uses functional representation organization
rather than architecture-specific layer identities.
\WFclear

\FloatBarrier
\section{Conclusion}
\label{sec_conclusion}
Machine unlearning verification requires internal evidence, not merely outputs
that resemble unlearning. ODR exposes this gap, while RUV rethinks the audit
boundary by tracing activation routes, feature states, neighborhoods, and
manifold support. Retained controls and joint calibration isolate
target-attributed responses from ordinary drift. Across methods, requests,
datasets, and architectures, RUV consistently verifies exact and approximate
unlearning executions and rejects output deception. Its response further tracks
request completion and rejects generic, wrong-target, and knowledge-preserving
target updates. These results establish forward representation auditing as a
practical basis for machine unlearning verification and motivate its extension
to generative and multimodal models.
\label{end:main}

\clearpage
\section*{AI Use Statement}
Generative AI tools assisted with manuscript language editing, LaTeX formatting,
translation, and code refinement. The authors are responsible for the final
manuscript and accompanying artifacts.
\section*{Reproducibility Statement}
The audit definitions, representation operators, and joint calibration procedure
are specified in Sections~\ref{sec:ruv_main} and~\ref{app:ruv}.
Their theoretical properties are established in Appendix~\ref{app:theory}.
Appendix~\ref{app:setup} details the data partitions, request manifests, model
pairing, hyperparameters, and evaluation protocol. Supplementary ODR and RUV
experiments are provided in Appendices~\ref{app:odr_results}
and~\ref{app:ruv_results}. An anonymized reference implementation and runnable
examples are available at
\url{https://anonymous.4open.science/r/iclr_veriun_core-E76F}.
\bibliographystyle{iclr2027_conference}
\bibliography{Veriun_ref}
\clearpage
\section*{Appendix Contents}
\pdfbookmark[0]{Appendix Contents}{appendixcontents}
\begingroup
\small
\setlength{\parskip}{0pt}
\linespread{0.94}\selectfont
\makeatletter
\@starttoc{atc}
\makeatother
\endgroup
\clearpage
\appendix
\raggedbottom
\addtocontents{atc}{\protect\appcontentsgroup{Background, Methods, and Theory}}
\appsection{Additional Related Work}
\label{app:related}

\appsubsection{Machine Unlearning Methods}

Machine unlearning methods are commonly divided by their guarantees into exact
and approximate unlearning. Exact unlearning requires the resulting model to
match, or be distributionally equivalent to, the result of applying the
prescribed training procedure to retained data. Full retraining provides the
direct reference, while dependency isolation, partitioned training, state
caching, and ensemble architectures reduce the portion of training that must be
recomputed
\citep{cao2015towards,ginart2019making,bourtoule2021machine,
brophy2021machine,yan2022arcane}. SISA, for example, confines the effect of a
request to affected shards and slices, then reconstructs the aggregate model
from components trained without the requested data
\citep{bourtoule2021machine}. Exact methods provide a clear counterfactual
target but inherit the storage, training, or architectural costs of maintaining
that reconstructable training structure.

Approximate unlearning relaxes retraining equivalence and modifies an existing
model at lower cost. Its mechanisms include, but are not limited to, retained
data fine tuning, gradient ascent or target relabeling, teacher--student
distillation, recorded update reversal, and selective parameter suppression
based on influence, Fisher information, or saliency
\citep{wu2020deltagrad,golatkar2020eternal,chundawat2023bad,
graves2021amnesiac,kurmanji2023scrub,fan2024salun,foster2024ssd}. These
strategies intervene at different scopes. Some replay or negate historical
updates, some optimize the full model against competing forget and retain
objectives, and others restrict changes to selected parameters. Consequently,
successful executions need not share a parameter trajectory, output
distribution, or layerwise response. This heterogeneity makes a verifier tied
to one update rule unsuitable as a general post hoc mechanism.

Independently of the guarantee, requests differ in their target content. A
class-level request removes all training samples associated with a class or
identity, whereas a sample-level request removes designated instances while
retaining other samples from the same class. Class removal can reorganize the
terminal decision geometry because an entire semantic group disappears from
training. Sample removal acts within shared class structure, so its effect is
typically more localized and distributed. Verification must therefore support
both heterogeneous update mechanisms and distinct response scales rather than
prescribe one universal signature.

\appsubsection{Verification of Machine Unlearning}

\paragraph{Process-bound evidence.}
Some verifiers arrange evidence before or during learning. Active methods
insert backdoor triggers, watermarks, canaries, model seeds, or influential
sample relations whose responses can later be queried
\citep{sommer2022athena,guo2024verifying,hu2022membership,gao2024verifi,
wang2026sms,xu2026really,zhang2025duplexguard,han2025vertical,
lu2025waterdrum}. Certified unlearning, replay, and proof systems instead bind
an audit to a prescribed optimizer, retained process records, or
provider-generated proof artifacts
\citep{chien2024certified,chien2024langevin,nguyen2022mcmc,
thudi2022necessity,weng2024proof,eisenhofer2025verifiable,
huynh2025certified,mu2025rewind}. These mechanisms connect a claim directly to
the learning process and can provide strong evidence under their intended
assumptions. They also require the verification arrangement to be anticipated,
the training algorithm to remain compatible, or the provider to retain and
expose the required artifacts. They are therefore most natural when
verification is designed into the model lifecycle, rather than introduced only
after an independently trained model receives an unlearning request.

\paragraph{Output-based post hoc evidence.}
Post hoc output methods audit an already submitted model through its prediction
interface. Forgotten-set accuracy measures the most direct task behavior;
membership inference compares forgotten samples with members or nonmembers;
confidence, entropy, and loss expose changes in predictive certainty; and
posterior-difference methods compare the responses of original and unlearned
models
\citep{shokri2017membership,lu2022fp2mia,carlini2022lira,
wang2025tape,wang2025lifecycle,jeon2026information,
vidal2025explainable}. These methods differ in statistic and calibration, but
observe the same boundary: the final response returned by the audited provider.
This common interface explains their broad deployability and their shared
exposure to response manipulation. A provider can calibrate confidence, smooth
posteriors, or return query-specific responses without performing the internal
update that the output is taken to represent. ODR makes this shared premise
explicit by constructing accepted output transcripts while introducing no
representation-level parameter update. The issue is therefore not confined to
one accuracy, membership, or posterior statistic; changing the statistic does
not change who controls its evidence source. This observation is also
consistent with prior evidence that verification can become fragile under
adaptive behavior by the audited party \citep{zhang2024fragile}.

\paragraph{Internal evidence and representation analysis.}
Internal verification asks whether the submitted model itself responds to the
request. TruVRF uses gradient sensitivity to measure how an unlearned model
responds to forgotten data \citep{zhou2025truvrf}. It moves beyond final
predictions, but requires white-box parameters and backward computation.
Moreover, a cross-entropy gradient contains the posterior residual
$p-e_y$, coupling the measured sensitivity to the model output used to
construct the audit loss. It therefore represents a substantially different
access and computation point from a forward representation audit.

Outside machine unlearning verification, representation analysis has developed
tools for identifying semantic units, comparing learned spaces, and
characterizing representation geometry
\citep{bau2017network,kornblith2019similarity,bengio2013representation,
belkin2001laplacian}. These studies establish that hidden states expose how a
model organizes information and how that organization changes across models.
They motivate moving an unlearning audit from the final prediction to the
forward computation that produces it. They do not, however, define a verifier
for a specified unlearning request. A raw activation, similarity, or distance
statistic does not distinguish a response concentrated on $D_u$ from ordinary
model-wide drift, nor does one statistic provide a universal signature for the
heterogeneous unlearning mechanisms described above.

\paragraph{From representation observation to verification.}
Post hoc verification necessarily reasons from observable consequences because
an external verifier generally lacks both the causal contribution of $D_u$ and
an independently available counterfactual model trained without it. Existing
methods instantiate this operational evidence at different boundaries:
prediction auditing uses task behavior and posteriors, membership inference
uses membership signals, active methods use planted responses, and TruVRF uses
backward gradient sensitivity
\citep{shokri2017membership,wang2025tape,sommer2022athena,
gao2024verifi,zhou2025truvrf}. The central design question is therefore not
whether an observable statistic changes, but whether that change can be
attributed to the requested data and converted into a calibrated decision.

RUV performs this conversion at the representation boundary. First, it evaluates
the same forgotten and retained inputs in the original and unlearned models and
uses matched retained controls to separate a response concentrated on $D_u$
from model-wide drift. Second, it follows complementary stages of representation
formation rather than treating one distance as a universal signature. Third,
joint permutation calibration preserves dependence among the views and adapts
the decision boundary to the model, data, and request under audit. RUV is thus
not merely a descriptive comparison of hidden spaces: it turns forward
representation responses into target-attributed, corroborated, and
request-calibrated evidence. Unlike output statistics, this evidence cannot be
reproduced by replacing only the final response; unlike gradient sensitivity,
it requires no backward execution. RUV thereby provides an algorithm-agnostic
post hoc verifier between provider-controlled output transcripts and
parameter-level backward inspection.

% \clearpage

\appsection{ODR Implementation Details}
\label{app:odr}
Section~\ref{sec_odr} defines both ODR constructions. This appendix records
only the implementation choices needed to reproduce them; optimization and
response equations are not repeated.

\appsubsection{ODR-Head Reproducibility}
ODR-Head implements Eqs.~\eqref{eq:main_forge}--\eqref{eq:main_odr_loss} with a
detached copy of the original model as the teacher. The teacher posterior used
to form $\eta(x)$ is therefore fixed throughout optimization rather than
recomputed from the changing head. All backbone parameters, BatchNorm buffers,
and stochastic inference state are frozen, and the backbone remains in
evaluation mode. Only $\theta_{\mathrm{head}}$ receives gradients. Forgotten
and retained minibatches are sampled independently for the two loss terms.
Consequently, ODR-Head introduces no representation-level parameter update;
only its final projection is changed.

\begin{figure}
    \centering
    \includegraphics[width=0.68\linewidth]{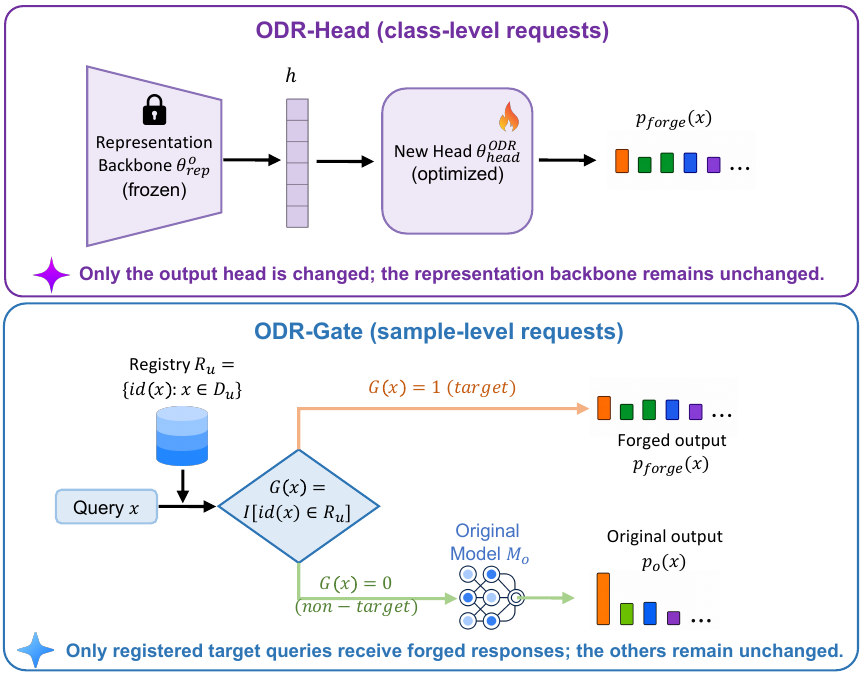}
    \caption{ODR-Head and ODR-Gate construct output-level unlearning responses
    without a representation-level parameter update.}
    \label{fig:odr_overview}
\end{figure}

\appsubsection{ODR-Gate Reproducibility}

ODR-Gate implements Eq.~\eqref{eq:main_gate} without modifying model
parameters. The request manifest supplies the immutable registry
$\mathcal R_u=\{\operatorname{id}(x):x\in D_u\}$. A single forward pass over
the held-out nonmember set constructs the class-indexed bank
\begin{equation}
\mathcal B_c=
\left\{p_o(x_j):(x_j,y_j)\in D_{\mathrm{aux}},\ y_j=c\right\}.
\end{equation}
Registered targets are sorted by stable identifier and assigned same-class
posteriors under a seed-controlled permutation. The assignment cycles through
$\mathcal B_{y_i}$ when a class contains more registered targets than available
nonmembers. This deterministic replay preserves the correctness, confidence,
entropy, and class competition of unseen same-class samples without requiring
a retrained oracle or an attack-specific hyperparameter.

Construction requires one forward pass over $D_{\mathrm{aux}}$ and one stored
posterior per registered target. Deployment adds one registry lookup and, for a
match, one posterior retrieval per query. These operations determine the costs
reported in Appendix~\ref{app:odr_results}.

\WFclear

\appsection{RUV Definitions and Calibration Details}
\label{app:ruv}

The main paper defines the four evidence views and final RUV decision. This
appendix specifies the architecture registry, metric implementation, cross-block
aggregation, and finite-sample calibration needed for reproduction.

\appsubsection{Representation Registry and Pooling}
\label{sec:ruv_chain}

Let $\mathcal L=(l_1,\ldots,l_{L-1},l_{\mathrm{pre}})$ contain the ordered
outputs of representation-bearing backbone blocks. A block is a functionally
distinct architecture-native module, such as a residual, convolutional, dense,
inverted residual, or Transformer encoder block. Primitive normalization,
activation, pooling, and classifier modules are not registered separately.
RUV uses $\mathcal L_{\mathrm{class}}=\{l_{\mathrm{pre}}\}$ and
$\mathcal L_{\mathrm{sample}}=\mathcal L\setminus\{l_{\mathrm{pre}}\}$, as in
Eq.~\eqref{eq:main_layers}.

For raw block output $h^{(l)}(x;\mathcal M)$, the common audit vector is
\begin{equation}
z^{(l)}(x;\mathcal M)=
\operatorname{vec}\!\left(\mathcal P_l[h^{(l)}(x;\mathcal M)]\right),
\label{eq:representation_vectorization}
\end{equation}
where $\mathcal P_l$ is fixed before auditing and shared by $\mathcal M_o$ and
$\mathcal M_u$. CNN blocks are assigned to $S$ architecture-native spatial
stages in forward order. If block $l$ belongs to stage $s(l)$ and has spatial
size $H_l\times W_l$, pooling is
\begin{equation}
\begin{aligned}
\bar g_{s(l)}&=
\begin{cases}
4,&s(l)\leq\lceil S/2\rceil,\ s(l)<S,\\
2,&\lceil S/2\rceil<s(l)<S,\\
1,&s(l)=S,
\end{cases}\\
g_l&=\min\{\bar g_{s(l)},H_l,W_l\},\qquad
\mathcal P_l[h]=\operatorname{AdaptiveAvgPool}_{g_l\times g_l}(h).
\end{aligned}
\label{eq:cnn_pooling}
\end{equation}
This retains coarse spatial organization in early stages without upsampling.
Transformer blocks use the designated summary token, and
$l_{\mathrm{pre}}$ uses the native classifier input.

Vector geometry uses cosine distance
\begin{equation}
d_{\cos}(a,b)=1-
\frac{a^\top b}
{\max(\|a\|_2,\epsilon_d)\max(\|b\|_2,\epsilon_d)},
\end{equation}
whereas activation-coordinate and neighbor-identity comparisons use Jaccard
overlap. The different metrics reflect different mathematical objects rather
than alternative choices for one statistic.

\appsubsection{Evidence Implementation and Cross-block Aggregation}

Every evidence view is computed per sample before group averaging. ARS and RS
are computed on corresponding coordinates of registered block pairs.

\textbf{ARS.}
For block $l$, define the dominant and thresholded activation sets
\begin{equation}
\begin{aligned}
A_q^{(l)}(x;\mathcal M)
&=\operatorname{Top}_{q_{\mathrm{top}}}
\left(|z^{(l)}(x;\mathcal M)|\right),\\
B_\gamma^{(l)}(x;\mathcal M)
&=\{m:z_m^{(l)}(x;\mathcal M)>\gamma_{\mathrm{act}}\}.
\end{aligned}
\end{equation}
Jaccard similarity is $J(A,B)=|A\cap B|/|A\cup B|$, with
$J(\varnothing,\varnothing)=1$. Substitution of the original and unlearned sets
into Eq.~\eqref{eq:main_ars} yields the ARS score. The weight $\omega_b$ balances
threshold crossings against changes among dominant coordinates.

\textbf{RS.}
RS is evaluated directly by Eq.~\eqref{eq:main_rs}. Independent vector
normalization makes the implementation identical to the cosine expression in
the main paper and prevents block scale from dominating the displacement.

\textbf{NPG.}
For reference representations
$\mathcal Z_{\mathrm{ref}}^{(l)}(\mathcal M)=
\{z^{(l)}(x';\mathcal M):x'\in D_{\mathrm{ref}}\}$, define
\begin{equation}
\mathcal N_{k_{\mathrm{NPG}}}^{(l)}(x;\mathcal M,D_{\mathrm{ref}})
=\operatorname{kNN}_{k_{\mathrm{NPG}}}^{d_{\cos}}
\left(z^{(l)}(x;\mathcal M),
\mathcal Z_{\mathrm{ref}}^{(l)}(\mathcal M)\right).
\end{equation}
Equation~\eqref{eq:main_npg} compares the identities in the original and
unlearned neighbor sets. Reference identities stay fixed while their
representations are recomputed; retrieval distances themselves are not
averaged.

\textbf{MRD.}
Let $D_{\mathrm{ref}}^y$ contain label-compatible reference samples and let
$\mathcal R_k^{(l)}(x;\mathcal M)$ be the $k_{\mathrm{MRD}}$ nearest elements of
that set under $d_{\cos}$. Their mean affinity is
\begin{equation}
\rho_k^{(l)}(x;\mathcal M)=
\frac{1}{|\mathcal R_k^{(l)}(x;\mathcal M)|}
\sum_{x'\in\mathcal R_k^{(l)}(x;\mathcal M)}
\operatorname{sim}_{\cos}
\left(z^{(l)}(x;\mathcal M),z^{(l)}(x';\mathcal M)\right).
\end{equation}
Raw affinity is calibrated against
$\mathcal C_k^{(l)}(\mathcal M)=
\{\rho_k^{(l)}(x';\mathcal M):x'\in D_{\mathrm{ref}}\}$, excluding self
matches. The percentile rank is
\begin{equation}
r_k^{(l)}(x;\mathcal M)=
\frac{1}{|\mathcal C_k^{(l)}(\mathcal M)|}
\sum_{c\in\mathcal C_k^{(l)}(\mathcal M)}
\mathbb I[c\leq\rho_k^{(l)}(x;\mathcal M)].
\end{equation}
Equation~\eqref{eq:main_mrd} takes the rank drop between models. Sample audits
use retained references; class audits use class-compatible held-out references
because the target class is absent from $D_r$. This distinguishes relative
support loss from the identity change measured by NPG
\citep{jiang2018trust}.

Block scores have architecture-dependent scales. RUV therefore standardizes
each block over the pooled, group-label-free audit set
$\mathcal X=D_u\cup D_r'$ before aggregation:
\begin{equation}
\begin{aligned}
\mu_{j,l}^{\mathrm{pool}}
&=\frac{1}{|\mathcal X|}\sum_{x\in\mathcal X}s_j^{(l)}(x),\\
\sigma_{j,l}^{\mathrm{pool}}
&=\sqrt{\frac{1}{|\mathcal X|}\sum_{x\in\mathcal X}
\left(s_j^{(l)}(x)-\mu_{j,l}^{\mathrm{pool}}\right)^2},\\
S_j^{(g)}(x)
&=\frac{1}{|\mathcal L_g|}\sum_{l\in\mathcal L_g}
\frac{s_j^{(l)}(x)-\mu_{j,l}^{\mathrm{pool}}}
{\sigma_{j,l}^{\mathrm{pool}}+\epsilon_s}.
\end{aligned}
\label{eq:cross_block_aggregation}
\end{equation}
Because the normalization does not use forgotten or retained group labels, it
cannot itself create group separation. We write $S_j(x)=S_j^{(g)}(x)$ for the
request granularity under audit.

\appsubsection{Target-specific Evidence and Joint Calibration}
\label{sec:ruv_decision}

For view $j$, the observed target-specific gap is
\begin{equation}
G_j^{\mathrm{obs}}=
\frac{1}{|D_u|}\sum_{x\in D_u}S_j(x)-
\frac{1}{|D_r'|}\sum_{x\in D_r'}S_j(x).
\label{eq:target_specific_gap}
\end{equation}
All views are oriented so that larger values indicate a stronger unlearning
response. Joint permutation estimates how large this gap can become under a
request-specific null. For permutation $b$, audit labels are reassigned while
group sizes and matching strata are preserved, producing pseudo groups
$\widetilde D_u^{(b)}$ and $\widetilde D_r^{(b)}$ and gap
\begin{equation}
G_j^{(b)}=
\frac{1}{|\widetilde D_u^{(b)}|}
\sum_{x\in\widetilde D_u^{(b)}}S_j(x)-
\frac{1}{|\widetilde D_r^{(b)}|}
\sum_{x\in\widetilde D_r^{(b)}}S_j(x).
\end{equation}
The samples, task labels, outputs, and four-view vectors remain fixed. The same
reassignment is applied to every view, paired model representations and all
augmented views remain in one sample block, sample audits are stratified by
class, and class audits use size-preserving pseudo-target assignments.

From $B$ permutations, define
\begin{equation}
\begin{aligned}
\mu_j^{(0)}&=\frac{1}{B}\sum_{b=1}^{B}G_j^{(b)},\\
\sigma_j^{(0)}&=
\sqrt{\frac{1}{B-1}\sum_{b=1}^{B}
\left(G_j^{(b)}-\mu_j^{(0)}\right)^2},\\
Z_j^{(t)}&=
\begin{cases}
(G_j^{(t)}-\mu_j^{(0)})/\sigma_j^{(0)},&\sigma_j^{(0)}>\epsilon_s,\\
0,&\text{otherwise},
\end{cases}
\quad t\in\{\mathrm{obs},1,\ldots,B\}.
\end{aligned}
\end{equation}
A degenerate null contributes zero evidence. Applying the second-largest rule
to the four standardized views gives $S_{\mathrm{RUV}}^{(t)}$ as in
Eq.~\eqref{eq:main_decision}. Its finite-sample fluctuation boundary is
\begin{equation}
\tau_\alpha=
Q_{1-\alpha}\!\left(
S_{\mathrm{RUV}}^{(1)},\ldots,S_{\mathrm{RUV}}^{(B)}
\right),
\label{eq:ruv_threshold}
\end{equation}
where $Q_{1-\alpha}$ uses order rank
$\min\{B,\lceil(B+1)(1-\alpha)\rceil\}$. The calibrated margin and decision are
\begin{equation}
\Delta_{\mathrm{RUV}}=
S_{\mathrm{RUV}}^{\mathrm{obs}}-\tau_\alpha,
\qquad
E_{\mathrm{RUV}}=\mathbb I[\Delta_{\mathrm{RUV}}>0].
\label{eq:ruv_margin}
\end{equation}
Using common permutations preserves empirical dependence among the four views
\citep{romano2005exact}. Across repeated trials, each trial is first judged
against its own boundary; reported verification rates average these binary
decisions rather than comparing an average score with an average threshold.

\appsubsection{Resistance and Cost of Internal Forgery}
\label{sec:internal_forgery}

RUV is executed by the verifier after the provider submits a model. The
matched retained subset, reference subset, audit randomization, per-view
evidence, and request-specific decision boundary remain verifier-side and are
not exposed before submission. Unlike an output verifier, RUV therefore
provides neither a compliant response that can be replayed nor a target
unlearned representation toward which the submitted model can be directly
aligned. Moving from output substitution to internal forgery changes the
attack from replacing one response to modifying the submitted model itself.

\begin{proposition}[Exclusion of representation-preserving submissions]
\label{prop:representation_preserving_exclusion}
Suppose a submitted strategy induces no target-specific response in the
audited representation blocks, so the joint per-sample evidence vectors are
exchangeable between $D_u$ and $D_r'$ under the matched audit design. Then
\begin{equation}
\Pr[E_{\mathrm{RUV}}=1\mid H_0]\leq\alpha.
\label{eq:representation_preserving_rejection}
\end{equation}
In particular, an output-only strategy that introduces no representation-level
parameter update remains in this null class under the matched design.
\end{proposition}

\begin{proof}
The strategy supplies no request-attributed component $t_j(x)$ in
Eq.~\eqref{eq:control_decomposition}. Under conditional exchangeability, its
observed assignment follows the joint permutation null, and
Theorem~\ref{thm:permutation_control} gives
Eq.~\eqref{eq:representation_preserving_rejection}. The result applies to the
composed RUV statistic and therefore accounts for dependence among the four
views.
\end{proof}

Let $a$ denote a provider strategy, $\mathcal M_a$ its submitted model,
$U(a)$ its retained utility, and $C(a)$ its implementation cost. For a fixed
audit realization, define the utility-preserving acceptance sets
\begin{equation}
\begin{aligned}
\mathcal A_{\mathrm{out}}
&=\{a:V_{\mathrm{out}}(a)=1,\ U(a)\geq u_{\min}\},\\
\mathcal A_{\mathrm{RUV}}
&=\{a:S_{\mathrm{RUV}}^{\mathrm{obs}}(a)>
\tau_\alpha(a),\ U(a)\geq u_{\min}\},\\
\mathcal A_{\mathrm{joint}}
&=\mathcal A_{\mathrm{out}}\cap\mathcal A_{\mathrm{RUV}}
\subseteq\mathcal A_{\mathrm{out}}.
\end{aligned}
\label{eq:forgery_acceptance_sets}
\end{equation}
Consequently,
\begin{equation}
\inf_{a\in\mathcal A_{\mathrm{joint}}}C(a)
\geq
\inf_{a\in\mathcal A_{\mathrm{out}}}C(a).
\label{eq:forgery_cost_relation}
\end{equation}
The added constraint is operationally substantial. Output deception changes
one exposed response while leaving the model untouched. Passing RUV additionally
requires the submitted model to produce a response concentrated on $D_u$
relative to $D_r'$ in at least two calibrated views. The model modification
must coordinate activation routes, feature states, neighbor identities, and
manifold support while preserving retained utility and the target information
that the provider intends to conceal.

The direct way to obtain a reliable internal target is to first produce a
compliant unlearned reference $\mathcal M_u^\star$ and align the submitted
model with its representations. If $C_{\mathrm{unl}}$ is the cost of that
execution, then
\begin{equation}
C_{\mathrm{ref}}(a)
=C_{\mathrm{unl}}+C_{\mathrm{align}}(a)+C_{\mathrm{keep}}(a)
\geq C_{\mathrm{unl}},
\label{eq:reference_forgery_cost}
\end{equation}
where $C_{\mathrm{align}}$ coordinates the submitted model with the reference
evidence and $C_{\mathrm{keep}}$ preserves the information removed by the
reference execution. Constructing the target therefore already incurs the
cost of unlearning, after which additional model-level optimization is needed.

RUV thus removes the low-cost interface shortcut exploited by ODR. A provider
cannot pass by substituting a posterior; it must coordinate target-attributed
changes across the model's internal computation. Reference-guided construction
offers no economic advantage over submitting the unlearned reference itself.

\appsection{Theoretical Analysis of ODR and RUV}
\label{app:theory}

This section answers two questions central to our design. First, why can ODR
systematically pass output-based verification without changing the internal
representation state? Second, why can RUV identify unlearning from forward
representations despite differences among unlearning algorithms? The first
answer follows from output-transcript indistinguishability. The second follows
from target attribution, multi-view corroboration, request-adaptive calibration,
and the detectability of a corroborated response. We use the evidence set
$\mathcal E=\{\mathrm{ARS},\mathrm{RS},\mathrm{NPG},\mathrm{MRD}\}$ and retain
the definitions of $S_j(x)$, $G_j^{\mathrm{obs}}$, $Z_j^{(t)}$,
$S_{\mathrm{RUV}}^{(t)}$, $\tau_\alpha$, and $E_{\mathrm{RUV}}$ from
Appendix~\ref{app:ruv}.

\appsubsection{Why ODR Works: Output-Transcript Indistinguishability}

The first question is whether changing the output statistic can resolve a
forged prediction transcript. The following bound shows that decisions
computed from the same transcript inherit its indistinguishability.
Consider an output-only verifier $V_{\mathrm{out}}$ that issues $T$ adaptive
queries. Its query $X_t$ may depend on internal randomness and on the history
$H_{t-1}=((X_1,Y_1),\ldots,(X_{t-1},Y_{t-1}))$, where
$Y_t=p_{\mathcal M}(X_t)$ is the response returned by model service
$\mathcal M$. Include the verifier's random seed $R$ in its complete view:
\begin{equation}
\mathsf T_T(\mathcal M)
=
\big(R,(X_1,Y_1),\ldots,(X_T,Y_T)\big).
\label{eq:output_transcript}
\end{equation}
Let $P_u^{\mathsf T}$ and $P_{\mathrm{ODR}}^{\mathsf T}$ be the transcript
laws induced by an unlearned model $\mathcal M_u$ and an ODR service
$\mathcal M_{\mathrm{ODR}}$, including all verifier and service randomness.
For a possibly randomized verifier, define its acceptance kernel
\begin{equation}
v(\mathbf t)
=
\Pr[V_{\mathrm{out}}(\mathbf t)=1\mid \mathsf T_T=\mathbf t],
\qquad 0\leq v(\mathbf t)\leq 1.
\label{eq:verifier_kernel}
\end{equation}

\begin{proposition}[Output transcript bound]
\label{prop:output_tv}
For every output-only verifier $V_{\mathrm{out}}$,
\begin{equation}
\begin{aligned}
\operatorname{Adv}_{\mathrm{out}}
&=
\left|
\int v(\mathbf t)\,\mathrm dP_u^{\mathsf T}(\mathbf t)
-
\int v(\mathbf t)\,\mathrm dP_{\mathrm{ODR}}^{\mathsf T}(\mathbf t)
\right|\\
&\leq
\sup_{0\leq f\leq 1}
\left|
\int f\,\mathrm dP_u^{\mathsf T}
-
\int f\,\mathrm dP_{\mathrm{ODR}}^{\mathsf T}
\right|\\
&=
\operatorname{TV}
\left(P_u^{\mathsf T},P_{\mathrm{ODR}}^{\mathsf T}\right).
\end{aligned}
\label{eq:output_tv_bound}
\end{equation}
Thus, identical transcript laws imply
$\operatorname{Adv}_{\mathrm{out}}=0$, while an $\varepsilon$ total-variation
match limits the advantage to at most $\varepsilon$.
\end{proposition}

\begin{proof}
Including $R$ makes the verifier's decision a common measurable function of
its complete view under both services. Any additional independent decision
randomness can be marginalized into the bounded function $v$ in
Eq.~\eqref{eq:verifier_kernel}. Its difference in acceptance
probability is therefore the first line of Eq.~\eqref{eq:output_tv_bound}.
Because $v$ belongs to the class $\{f:0\leq f\leq1\}$ over which total
variation takes the supremum, the second line follows immediately. For a
deterministic verifier, $v=\mathbb I_{\mathcal A}$ for its acceptance region
$\mathcal A$, recovering
$|P_u^{\mathsf T}(\mathcal A)-P_{\mathrm{ODR}}^{\mathsf T}(\mathcal A)|$.
\end{proof}

The bound also quantifies how local response matching accumulates over an
adaptive interaction. Let
$Q_{u,t}(\cdot\mid h_{t-1},x_t)$ and
$Q_{\mathrm{ODR},t}(\cdot\mid h_{t-1},x_t)$ denote the two conditional response
laws at round $t$, conditional also on the common verifier seed $R$
(suppressed in the notation).

\begin{corollary}[Adaptive composition of response mismatch]
\label{cor:adaptive_composition}
Suppose that, for every common seed, history, and query, with
$0\leq\varepsilon_t\leq1$,
\begin{equation}
\operatorname{TV}\!\left(
Q_{u,t}(\cdot\mid h_{t-1},x_t),
Q_{\mathrm{ODR},t}(\cdot\mid h_{t-1},x_t)
\right)
\leq \varepsilon_t.
\label{eq:conditional_tv}
\end{equation}
Then
\begin{equation}
\operatorname{Adv}_{\mathrm{out}}
\leq
\operatorname{TV}(P_u^{\mathsf T},P_{\mathrm{ODR}}^{\mathsf T})
\leq
1-\prod_{t=1}^{T}(1-\varepsilon_t)
\leq
\sum_{t=1}^{T}\varepsilon_t.
\label{eq:adaptive_tv_composition}
\end{equation}
\end{corollary}

\begin{proof}
Couple the two interactions with the same verifier randomness. As long as the
histories agree, the verifier issues the same next query. A maximal coupling
of the two conditional response laws then keeps the round-$t$ responses equal
with probability at least $1-\varepsilon_t$. Hence the probability that the
entire transcripts remain equal is at least
$\prod_{t=1}^{T}(1-\varepsilon_t)$. The coupling characterization of total
variation yields the middle inequality in
Eq.~\eqref{eq:adaptive_tv_composition}; the final inequality follows from
$1-\prod_t(1-\varepsilon_t)\leq\sum_t\varepsilon_t$.
\end{proof}

For any statistic $\phi$ computed only from the transcript, total variation
also obeys the data-processing inequality
\begin{equation}
\operatorname{TV}
\left(
\mathcal L(\phi(\mathsf T_T(\mathcal M_u))),
\mathcal L(\phi(\mathsf T_T(\mathcal M_{\mathrm{ODR}})))
\right)
\leq
\operatorname{TV}(P_u^{\mathsf T},P_{\mathrm{ODR}}^{\mathsf T}).
\label{eq:output_data_processing}
\end{equation}
Accuracy, confidence, membership scores, and posterior differences are all such
statistics. ODR therefore does not need a separate attack for every output
verifier. It controls the common evidence source from which these criteria are
computed. Once the exposed transcript matches the behavior accepted as
unlearning, no post-processing of that transcript can recover whether the
internal representation state was updated. This establishes why ODR can
simultaneously deceive output criteria built from different statistics.

\appsubsection{Why RUV Works: Target Attribution}

The next question is which part of a representation response is concentrated
on the request. A model update can also move retained samples, so absolute
change alone does not isolate the requested effect. Each RUV view produces a
per-sample score $S_j(x)$; decompose it as
\begin{equation}
S_j(x)
=
c_j+b_j(x)+\mathbb I[x\in D_u]t_j(x)+\varepsilon_j(x),
\qquad j\in\mathcal E,
\label{eq:control_decomposition}
\end{equation}
where $c_j$ is common model drift, $b_j(x)$ is sample-dependent nuisance
drift, $t_j(x)$ is the request-attributed response, and $\varepsilon_j(x)$ is
zero-mean sampling noise. Enumerate
$D_u=\{x_{u,1},\ldots,x_{u,n}\}$ and
$D_r'=\{x_{r,1},\ldots,x_{r,n}\}$. Substituting
Eq.~\eqref{eq:control_decomposition} into the observed gap gives the exact
finite-sample identity
\begin{equation}
\begin{aligned}
G_j^{\mathrm{obs}}
&=
\frac1n\sum_{i=1}^{n}S_j(x_{u,i})
-
\frac1n\sum_{i=1}^{n}S_j(x_{r,i})\\
&=
\underbrace{\frac1n\sum_{i=1}^{n}t_j(x_{u,i})}_{\overline t_{u,j}}
\\
&\quad+
\underbrace{\left[
\frac1n\sum_{i=1}^{n}b_j(x_{u,i})
-
\frac1n\sum_{i=1}^{n}b_j(x_{r,i})
\right]}_{\overline b_{u,j}-\overline b_{r,j}}
\\
&\quad+
\underbrace{\left[
\frac1n\sum_{i=1}^{n}\varepsilon_j(x_{u,i})
-
\frac1n\sum_{i=1}^{n}\varepsilon_j(x_{r,i})
\right]}_{\overline\varepsilon_{u,j}-\overline\varepsilon_{r,j}}.
\end{aligned}
\label{eq:gap_expansion}
\end{equation}
The common component $c_j$ disappears algebraically before any distributional
assumption is invoked.

\begin{proposition}[Unbiased target attribution]
\label{prop:control_cancellation}
Suppose $|D_u|=|D_r'|=n$,
\begin{equation}
\mathbb E[\varepsilon_j(x)\mid x\in D_u]
=
\mathbb E[\varepsilon_j(x)\mid x\in D_r']=0,
\qquad
\mathbb E_{D_u}[b_j(x)]=\mathbb E_{D_r'}[b_j(x)].
\label{eq:nuisance_matching}
\end{equation}
Then
\begin{equation}
\mathbb E[G_j^{\mathrm{obs}}]
=
\mathbb E_{x\sim D_u}[t_j(x)]
\equiv \delta_j.
\label{eq:gap_expectation}
\end{equation}
Under the null response $t_j(x)=0$, the expected gap is zero.
\end{proposition}

\begin{proof}
Taking expectations on both sides of Eq.~\eqref{eq:gap_expansion} yields
\begin{equation}
\begin{aligned}
\mathbb E[G_j^{\mathrm{obs}}]
&=
\mathbb E_{D_u}[t_j(x)]
+\mathbb E_{D_u}[b_j(x)]-\mathbb E_{D_r'}[b_j(x)]\\
&\quad+
\mathbb E[\varepsilon_j(x)\mid x\in D_u]
-\mathbb E[\varepsilon_j(x)\mid x\in D_r'].
\end{aligned}
\label{eq:gap_expectation_expanded}
\end{equation}
The second line and the nuisance difference vanish by
Eq.~\eqref{eq:nuisance_matching}, leaving $\delta_j$.
\end{proof}

The mean identity explains attribution; its uncertainty determines how much
evidence an audit needs. Under an independent-sampling model for the scores,
the same decomposition provides a finite-sample fluctuation bound. Define
\begin{equation}
R_{u,j}(x)=b_j(x)+t_j(x)+\varepsilon_j(x),
\qquad
R_{r,j}(x)=b_j(x)+\varepsilon_j(x),
\label{eq:group_residuals}
\end{equation}
with variances $\sigma_{u,j}^2$ and $\sigma_{r,j}^2$.

\begin{proposition}[Variance and concentration of the attributed gap]
\label{prop:gap_concentration}
Under Eq.~\eqref{eq:nuisance_matching}, if observations are identically
distributed and independent within each group and the two sampled group
means are independent, then
\begin{equation}
\operatorname{Var}(G_j^{\mathrm{obs}})
=
\frac{\sigma_{u,j}^2+\sigma_{r,j}^2}{n}.
\label{eq:gap_variance}
\end{equation}
Consequently, for every $\eta>0$,
\begin{equation}
\Pr\!\left(
|G_j^{\mathrm{obs}}-\delta_j|\geq\eta
\right)
\leq
\frac{\sigma_{u,j}^2+\sigma_{r,j}^2}{n\eta^2}.
\label{eq:gap_chebyshev}
\end{equation}
For paired controls with covariance
$\kappa_j=\operatorname{Cov}(\overline R_{u,j},\overline R_{r,j})$,
the right-hand variance becomes
$(\sigma_{u,j}^2+\sigma_{r,j}^2)/n-2\kappa_j$.
\end{proposition}

\begin{proof}
Since $c_j$ is constant and cancels,
$G_j^{\mathrm{obs}}=\overline R_{u,j}-\overline R_{r,j}$. Hence
\begin{equation}
\begin{aligned}
\operatorname{Var}(G_j^{\mathrm{obs}})
&=
\operatorname{Var}(\overline R_{u,j})
+\operatorname{Var}(\overline R_{r,j})
-2\operatorname{Cov}(\overline R_{u,j},\overline R_{r,j})\\
&=
\frac{\sigma_{u,j}^2}{n}
+\frac{\sigma_{r,j}^2}{n}
-2\kappa_j.
\end{aligned}
\label{eq:gap_variance_derivation}
\end{equation}
Setting $\kappa_j=0$ gives Eq.~\eqref{eq:gap_variance}; applying Chebyshev's
inequality around the mean in Eq.~\eqref{eq:gap_expectation} gives
Eq.~\eqref{eq:gap_chebyshev}.
\end{proof}

Equations~\eqref{eq:gap_expectation}--\eqref{eq:gap_chebyshev} separate two
roles of the control design. Equal expected nuisance responses isolate the
target effect; under the stated sampling assumptions, increasing the audit set
contracts uncertainty at rate $n^{-1/2}$. Class matching helps balance nuisance
responses in sample audits, but equal group sizes alone do not establish this
balance in class audits. Shared references and pooled normalization can also
couple scores; the first equality in
Eq.~\eqref{eq:gap_variance_derivation} remains valid, with within-group
covariances included in each mean's variance. The next analysis calibrates the joint score conditional on the
observed score vectors.

\appsubsection{Why RUV Works: Corroboration and Calibration}

This part answers two linked questions: how to require corroboration without
requiring every view to respond, and how to calibrate that requirement when
the views are dependent. Write the four standardized effects in descending order as
$Z_{(1)}\geq Z_{(2)}\geq Z_{(3)}\geq Z_{(4)}$. RUV uses
$S_{\mathrm{RUV}}=Z_{(2)}$.

\begin{lemma}[Two-view corroboration]
\label{lem:second_largest}
For any $\mathbf z=(z_j)_{j\in\mathcal E}\in\mathbb R^4$ and threshold $c$,
\begin{equation}
\operatorname{2ndLargest}_{j\in\mathcal E}z_j>c
\quad\Longleftrightarrow\quad
\sum_{j\in\mathcal E}\mathbb I[z_j>c]\geq2.
\label{eq:second_largest_equivalence}
\end{equation}
\end{lemma}

\begin{proof}
Let $z_{(1)}\geq z_{(2)}\geq z_{(3)}\geq z_{(4)}$. Then
\begin{equation}
z_{(2)}>c
\iff
z_{(1)}>c\ \text{and}\ z_{(2)}>c
\iff
\sum_{j\in\mathcal E}\mathbb I[z_j>c]\geq2.
\end{equation}
Thus one extreme view cannot determine the joint decision.
\end{proof}

Let $\Pi$ be the set of admissible audit-label assignments preserving group
sizes and all matching strata, and let $N_\Pi=|\Pi|$. For
$\pi\in\Pi$, recompute the four gaps jointly and define
\begin{equation}
\begin{aligned}
\mu_j^\Pi
&=\frac1{N_\Pi}\sum_{\pi\in\Pi}G_j(\pi),\\
(\sigma_j^\Pi)^2
&=\frac1{N_\Pi}\sum_{\pi\in\Pi}
\left(G_j(\pi)-\mu_j^\Pi\right)^2,\\
Z_j(\pi)
&=\frac{G_j(\pi)-\mu_j^\Pi}{\sigma_j^\Pi},\\
T(\pi)
&=\operatorname{2ndLargest}_{j\in\mathcal E}Z_j(\pi).
\end{aligned}
\label{eq:ideal_permutation_statistics}
\end{equation}
Degenerate views with $\sigma_j^\Pi\leq\epsilon_s$ are assigned zero, as in
the implemented rule. Let
$T_{[1]}\leq\cdots\leq T_{[N_\Pi]}$ be the ordered ideal scores and set
\begin{equation}
k_\alpha=\left\lceil(1-\alpha)N_\Pi\right\rceil,
\qquad
\tau_\alpha^\Pi=T_{[k_\alpha]}.
\label{eq:ideal_permutation_quantile}
\end{equation}

\begin{theorem}[Conditional false-positive control]
\label{thm:permutation_control}
Under the null hypothesis that audit labels are exchangeable over $\Pi$
conditional on the pooled score vectors and matching strata,
\begin{equation}
\Pr_{H_0}\!\left[
S_{\mathrm{RUV}}^{\mathrm{obs}}>\tau_\alpha^\Pi
\,\middle|\,
\{S_j(x)\}_{x\in\mathcal X,j\in\mathcal E}
\right]
\leq\alpha.
\label{eq:permutation_level}
\end{equation}
This result permits arbitrary dependence among the four views.
\end{theorem}

\begin{proof}
Conditional exchangeability makes the observed assignment $\pi_0$ uniform on
$\Pi$. Therefore
\begin{equation}
\begin{aligned}
\Pr_{H_0}[T(\pi_0)>\tau_\alpha^\Pi\mid\mathcal X]
&=
\frac1{N_\Pi}
\sum_{\pi\in\Pi}
\mathbb I[T(\pi)>T_{[k_\alpha]}]\\
&\leq
\frac{N_\Pi-k_\alpha}{N_\Pi}
\leq\alpha.
\end{aligned}
\label{eq:permutation_rank_proof}
\end{equation}
The same $\pi$ acts on the complete four-dimensional score vector of every
sample. Hence Eq.~\eqref{eq:permutation_rank_proof} calibrates the composed
statistic $T$ directly and never assumes independence among ARS, RS, NPG, and
MRD \citep{romano2005exact}.
\end{proof}

The theorem explains the role of joint permutation: it calibrates the complete
decision rather than four separate tests. Its exchangeability condition concerns
the joint score distribution, which is stronger than equal expected nuisance
responses in Eq.~\eqref{eq:nuisance_matching}. Size-preserving assignments
define the class-audit reference distribution; nominal false-positive control
requires this same condition. The implementation approximates the reference
distribution with $B$ assignments from $\Pi$. Let
\begin{equation}
\widehat F_B(s)
=
\frac1B\sum_{b=1}^{B}
\mathbb I[S_{\mathrm{RUV}}^{(b)}\leq s]
\label{eq:mc_empirical_cdf}
\end{equation}
be the empirical joint-null distribution.

\begin{proposition}[Consistency of Monte Carlo calibration]
\label{prop:mc_consistency}
Assume each ideal null standard deviation is separated from $\epsilon_s$,
with inactive views assigned zero, and the target quantile is
identifiable, namely
$F_\Pi((\tau_\alpha^\Pi)^-)<1-\alpha<F_\Pi(\tau_\alpha^\Pi)$.
For independent uniform draws
$\pi_1,\ldots,\pi_B$ from $\Pi$,
\begin{equation}
\begin{aligned}
\mu_j^{(0)}&\xrightarrow{\mathrm{a.s.}}\mu_j^\Pi,
&\sigma_j^{(0)}&\xrightarrow{\mathrm{a.s.}}\sigma_j^\Pi,\\
\widehat F_B(s)&\xrightarrow{\mathrm{a.s.}}F_\Pi(s)
\quad\text{at continuity points }s,
&\tau_\alpha&\xrightarrow{\mathrm{a.s.}}\tau_\alpha^\Pi.
\end{aligned}
\label{eq:mc_convergence}
\end{equation}
Consequently, the implemented decision converges to the ideal permutation
decision whenever the observed score is not on the limiting boundary.
\end{proposition}

\begin{proof}
The first two limits follow from the strong law applied to $G_j(\pi_b)$ and
$G_j(\pi_b)^2$. Separation from $\epsilon_s$ makes the active set eventually
constant. Since $\Pi$ is finite, the estimated standardized score converges
uniformly over its assignments to $T(\pi)$. The empirical frequencies of
assignments also converge, giving CDF convergence at continuity points.
The strict quantile inequalities place $1-\alpha$ inside one limiting atom;
the empirical quantile therefore converges to that atom's location. The
implemented order rank divided by $B$ tends to $1-\alpha$, so its finite-sample
rank correction has the same limit. The observed statistic converges by the
same continuity argument, establishing the decision claim away from ties.
\end{proof}

\appsubsection{Why RUV Works: Detectability}

Calibration determines the fluctuation boundary; detectability asks when a
response crosses it. This part relates the strength of two corroborating views
to audit size, explaining why distributed sample responses can require more
observations than stronger class responses. The asymptotic calculation uses
the ideal active-view rule with $\epsilon_s=0$; the finite-sample bound below
includes the positive numerical cutoff used in implementation.

\begin{theorem}[Consistency under corroborated response]
\label{thm:ruv_power}
For the ideal rule with $\epsilon_s=0$, let $n=|D_u|=|D_r'|$ increase under
the matched audit design. Suppose that for
some set $\mathcal J_\star\subseteq\mathcal E$ with
$|\mathcal J_\star|\geq2$,
\begin{equation}
G_j^{\mathrm{obs}}\xrightarrow{p}\delta_j>0,
\qquad
\mu_j^{(0)}=o_p(1),
\qquad
\sqrt n\,\sigma_j^{(0)}\xrightarrow{p}s_j\in(0,\infty),
\quad j\in\mathcal J_\star,
\label{eq:positive_effect}
\end{equation}
and $\tau_\alpha=O_p(1)$. Then
\begin{equation}
\frac{Z_j^{\mathrm{obs}}}{\sqrt n}
\xrightarrow{p}
\frac{\delta_j}{s_j}>0,
\quad j\in\mathcal J_\star,
\qquad
\Pr[E_{\mathrm{RUV}}=1]\longrightarrow1.
\label{eq:ruv_consistency}
\end{equation}
\end{theorem}

\begin{proof}
For every $j\in\mathcal J_\star$, rewrite the standardized effect as
\begin{equation}
\frac{Z_j^{\mathrm{obs}}}{\sqrt n}
=
\frac{G_j^{\mathrm{obs}}-\mu_j^{(0)}}
{\sqrt n\,\sigma_j^{(0)}}.
\label{eq:z_power_decomposition}
\end{equation}
The numerator converges in probability to $\delta_j$ and the denominator to
$s_j>0$. Slutsky's theorem gives the first limit in
Eq.~\eqref{eq:ruv_consistency}, so at least two standardized views diverge to
$+\infty$. By Lemma~\ref{lem:second_largest},
\begin{equation}
S_{\mathrm{RUV}}^{\mathrm{obs}}
=
\operatorname{2ndLargest}_{j\in\mathcal E}Z_j^{\mathrm{obs}}
\xrightarrow{p}+\infty.
\label{eq:aggregate_divergence}
\end{equation}
For any $M>0$,
\begin{equation}
\begin{aligned}
\Pr[E_{\mathrm{RUV}}=0]
&=
\Pr[S_{\mathrm{RUV}}^{\mathrm{obs}}\leq\tau_\alpha]\\
&\leq
\Pr[S_{\mathrm{RUV}}^{\mathrm{obs}}\leq M]
+\Pr[\tau_\alpha>M].
\end{aligned}
\label{eq:power_probability_bound}
\end{equation}
The first term tends to zero for fixed $M$, and tightness of
$\tau_\alpha=O_p(1)$ makes the second arbitrarily small for sufficiently large
$M$. Hence $\Pr[E_{\mathrm{RUV}}=0]\to0$.
\end{proof}

A finite-sample bound connects this signal-to-noise argument to the implemented
rule. Let two corroborating views satisfy
$\mathbb E[G_j^{\mathrm{obs}}]=\delta_j>0$ and
$\operatorname{Var}(G_j^{\mathrm{obs}})\leq v_j/n$, including any score
dependence in this variance bound. Define their calibrated gap thresholds as
$h_{j,n}=\mu_j^{(0)}+\tau_\alpha\sigma_j^{(0)}$ and the event
$\mathcal H_n=\{\sigma_j^{(0)}>\epsilon_s,\,
h_{j,n}\leq\delta_j/2:\,j\in\mathcal J_2\}$.
A union bound and Eq.~\eqref{eq:gap_chebyshev} give
\begin{equation}
\Pr[E_{\mathrm{RUV}}=0]
\leq
\Pr(\mathcal H_n^c)
+
\sum_{j\in\mathcal J_2}
\Pr[G_j^{\mathrm{obs}}\leq h_{j,n},\mathcal H_n]
\leq
\Pr(\mathcal H_n^c)
+
\sum_{j\in\mathcal J_2}
\frac{4v_j}{n\delta_j^2},
\qquad |\mathcal J_2|=2.
\label{eq:finite_sample_power}
\end{equation}
Equation~\eqref{eq:finite_sample_power} identifies the relevant quantities:
both views must be active and their calibrated thresholds must leave room for
the target effects. For comparable variances and boundary events, larger
$\delta_j$ reduces the required audit size through $n\delta_j^2$.
This explains the observed class/sample ordering when class requests produce
stronger effects; granularity itself is not an assumption of the theorem.

\addtocontents{atc}{\protect\appcontentsgroup{Reproducibility Details}}
\appsection{Experimental Details}\label{app:setup}

\appsubsection{Requests and Reporting Protocol}

Class requests remove all training samples from one class or identity. Sample
requests use class- or identity-stratified sampling without replacement, with
$\rho=10\%$ in the main experiments and
$\rho\in\{1\%,5\%,10\%,20\%\}$ in the scale study. Every request is stored in
an immutable manifest shared by all unlearning methods and verifiers.

For each dataset and request level, we train ten original models using
independent model seeds. Each model seed controls initialization, data order,
and training stochasticity. We then generate ten request seeds per original
model, each defining an independent immutable manifest. This produces 100 class
submissions and 100 sample submissions for every unlearning method and its
corresponding ODR construction. The same manifests are reused across all
methods and verifiers within a setting.

Each submission is audited under ten verifier seeds. These seeds resample the
matched retained control, reference subset, and joint permutations, but never
change the original model, unlearned model, or request manifest. Let
$d_{s,q,v}\in\{0,1\}$ denote the relevant correct-decision indicator for model
seed $s$, request seed $q$, and verifier seed $v$. We first average the ten
verifier outcomes for each submission and the ten requests for each original
model,
\begin{equation}
\bar d_{s,q}=\frac{1}{10}\sum_{v=1}^{10}d_{s,q,v},
\qquad
\bar d_s=\frac{1}{10}\sum_{q=1}^{10}\bar d_{s,q}.
\label{eq:seed_aggregation}
\end{equation}
Consequently, each reported setting contains
$N=10\times10\times10=1000$ binary audits, with success count and point estimate
\begin{equation}
K=\sum_{s=1}^{10}\sum_{q=1}^{10}\sum_{v=1}^{10}d_{s,q,v},
\qquad
\widehat p=\frac{K}{1000}=\frac{1}{10}\sum_{s=1}^{10}\bar d_s.
\label{eq:binary_audit_count}
\end{equation}
The three-decimal point estimates in the main tables therefore report the
corresponding success counts directly; for example, $0.961$ denotes
$961/1000$ successful binary audits. Standard error is computed across the ten
independent model-level summaries $\bar d_s$, not across all 1000 outcomes.
Controlled ablations use the same hierarchy but report their standard
deviation. The verifier repetitions therefore stabilize each submission
without being treated as independently trained models or requests.
When the same ODR audits are reported as deception success in
Table~\ref{tab:odr_dsr} and rejection in Table~\ref{tab:verification_main}, the
indicators satisfy $d_{s,q,v}^{\mathrm{reject}}=1-d_{s,q,v}^{\mathrm{DSR}}$.
Their point estimates are therefore complementary and their model-level
standard errors are identical.

For verification accuracy, $d_{s,q,v}=1$ denotes acceptance of a configured
unlearning execution or rejection of its adversarial submission. For ODR
deception success rate, it denotes acceptance of ODR by the evaluated output
verifier. Supplementary output analyses additionally report
$\operatorname{Acc}_u$, $\operatorname{Conf}_u$, $\operatorname{Acc}_r$, and
$\operatorname{Acc}_{\mathrm{test}}$ for forgotten behavior, confidence,
retained utility, and test generalization.

\appsubsection{Data and Model Configuration}

The main dataset-model pairs are CIFAR10/ResNet18, CIFAR100/ResNet34, and
FaceScrub/ResNet50 \citep{krizhevsky2009learning,ng2014facescrub,he2016resnet}.
CIFAR experiments use the official test sets and stratified validation splits.
FaceScrub is cleaned of unreadable and duplicate files, split by identity into
70:10:20 training, validation, and test partitions, cropped using the provided
regions, and resized to $112\times112$. Training uses random cropping and
horizontal flipping; auditing uses deterministic transforms.

Architecture robustness on CIFAR10 additionally evaluates ResNet34, ResNet50,
VGG16, DenseNet121, MobileNetV2, and ViT-S/16
\citep{simonyan2015vgg,huang2017densenet,sandler2018mobilenetv2,
dosovitskiy2021vit}. Forward hooks collect architecture-native block outputs.
Equation~\eqref{eq:cnn_pooling} defines CNN pooling, ViT blocks use the summary
token, and the preclassifier block uses its native classifier input. Class RUV
uses the preclassifier representation; sample RUV averages standardized block
scores over all nonterminal blocks. Controls are sampled without replacement,
with $|D_r'|=|D_u|$, and are class matched for sample requests.

\appsubsection{Unlearning and Verification Configurations}

Retraining and SISA provide exact baselines; Amnesiac Machine Learning, SCRUB,
SalUn, and SSD cover distinct approximate mechanisms
\citep{bourtoule2021machine,graves2021amnesiac,kurmanji2023scrub,
fan2024salun,foster2024ssd}. Retraining uses the original initialization and
trains from scratch on $D_r$. SISA uses five shards and five slices. Approximate
methods use released implementations with validation-selected settings fixed
before testing.

Accuracy, LiRA, TAPE, and TruVRF respectively provide task behavior, membership,
posterior-difference, and gradient-sensitivity baselines
\citep{carlini2022lira,wang2025tape,zhou2025truvrf}. Every verifier receives the
same model pair, partition, manifest, and request. TruVRF follows its original
model-sensitivity protocol: the verifier recomputes predictions from the
submitted model and extracts the resulting backward sensitivity through the
native computation graph. For ODR-Gate, the replayed response is therefore not
used in gradient extraction. Baseline calibration does not use test requests.

ODR-Head freezes the backbone state and trains the classifier for eight Adam
epochs with learning rate $10^{-3}$, $\lambda_{\mathrm{ODR}}=0.5$, and
perturbation standard deviation $0.01$; $\alpha_{\mathrm{ODR}}$ is selected on
validation data and fixed before testing. ODR-Gate uses exact manifest indices
and seed-controlled same-class posteriors collected from the held-out nonmember
split only for registered queries.

\appsubsection{RUV Settings}

RUV uses $q_{\mathrm{top}}=0.1$, $\omega_b=0.5$, and
$\gamma_{\mathrm{act}}=0$ for ARS, $k_{\mathrm{NPG}}=10$,
$k_{\mathrm{MRD}}=1$, $N_{\mathrm{ref}}\leq10{,}000$, $B=1000$, and
$\alpha=0.05$. Sample references come from retained data; class references use
class-compatible held-out samples.

\addtocontents{atc}{\protect\appcontentsgroup{Supplementary Experiments}}
\appsection{Supplementary ODR Experiments}
\label{app:odr_results}
\label{app:results}

\appsubsection{Prediction Behavior and Retained Utility}
\begin{table}[H]
\centering
\caption{Output behavior and retained utility across unlearning methods.}
\label{tab:odr_behavior}
\fontsize{7.5}{9}\selectfont
\setlength{\tabcolsep}{3pt}
\renewcommand{\arraystretch}{1.12}
\begin{tabular*}{\linewidth}{@{\extracolsep{\fill}}cccccccccc@{}}
\toprule
& & \multicolumn{4}{c}{Class level} & \multicolumn{4}{c}{Sample level} \\
\cmidrule(lr){3-6}\cmidrule(lr){7-10}
\multicolumn{1}{c}{Dataset} & \multicolumn{1}{c}{Submission} & $\mathrm{Acc}_u$ & $\mathrm{Conf}_u$ & $\mathrm{Acc}_r$ & $\mathrm{Acc}_{test}$ & $\mathrm{Acc}_u$ & $\mathrm{Conf}_u$ & $\mathrm{Acc}_r$ & $\mathrm{Acc}_{test}$ \\
\midrule
\multirow{8}{*}{CIFAR10} & Original & \est{0.992}{0.007} & \est{0.970}{0.007} & \est{0.993}{0.004} & \est{0.940}{0.007} & \est{0.989}{0.006} & \est{0.968}{0.009} & \est{0.992}{0.004} & \est{0.938}{0.006} \\
 & Retraining & \est{0.015}{0.012} & \est{0.053}{0.013} & \est{0.987}{0.005} & \est{0.838}{0.011} & \est{0.925}{0.009} & \est{0.872}{0.018} & \est{0.985}{0.005} & \est{0.924}{0.009} \\
 & SISA & \est{0.019}{0.015} & \est{0.065}{0.014} & \est{0.989}{0.005} & \est{0.840}{0.010} & \est{0.927}{0.008} & \est{0.875}{0.019} & \est{0.986}{0.005} & \est{0.926}{0.009} \\
 & Amnesiac & \est{0.110}{0.018} & \est{0.223}{0.012} & \est{0.972}{0.004} & \est{0.862}{0.009} & \est{0.842}{0.013} & \est{0.800}{0.019} & \est{0.972}{0.004} & \est{0.907}{0.005} \\
 & SCRUB & \est{0.073}{0.013} & \est{0.183}{0.016} & \est{0.979}{0.005} & \est{0.875}{0.006} & \est{0.882}{0.024} & \est{0.837}{0.029} & \est{0.979}{0.004} & \est{0.915}{0.010} \\
 & SalUn & \est{0.045}{0.011} & \est{0.134}{0.015} & \est{0.980}{0.002} & \est{0.876}{0.007} & \est{0.868}{0.013} & \est{0.823}{0.014} & \est{0.979}{0.004} & \est{0.914}{0.005} \\
 & SSD & \est{0.128}{0.016} & \est{0.257}{0.020} & \est{0.983}{0.003} & \est{0.889}{0.006} & \est{0.900}{0.008} & \est{0.856}{0.021} & \est{0.985}{0.006} & \est{0.926}{0.009} \\
 & \textbf{ODR} & \textbf{\est{0.086}{0.018}} & \textbf{\est{0.157}{0.021}} & \textbf{\est{0.982}{0.007}} & \textbf{\est{0.858}{0.011}} & \textbf{\est{0.936}{0.008}} & \textbf{\est{0.891}{0.015}} & \textbf{\est{0.990}{0.004}} & \textbf{\est{0.939}{0.006}} \\
\midrule
\multirow{8}{*}{CIFAR100} & Original & \est{0.977}{0.005} & \est{0.938}{0.006} & \est{0.982}{0.006} & \est{0.772}{0.008} & \est{0.980}{0.009} & \est{0.943}{0.014} & \est{0.980}{0.008} & \est{0.772}{0.007} \\
 & Retraining & \est{0.005}{0.009} & \est{0.016}{0.014} & \est{0.980}{0.006} & \est{0.761}{0.012} & \est{0.749}{0.017} & \est{0.689}{0.019} & \est{0.980}{0.006} & \est{0.761}{0.011} \\
 & SISA & \est{0.007}{0.011} & \est{0.019}{0.017} & \est{0.982}{0.007} & \est{0.764}{0.013} & \est{0.754}{0.016} & \est{0.694}{0.017} & \est{0.981}{0.005} & \est{0.764}{0.011} \\
 & Amnesiac & \est{0.032}{0.025} & \est{0.065}{0.023} & \est{0.963}{0.006} & \est{0.737}{0.013} & \est{0.640}{0.023} & \est{0.581}{0.021} & \est{0.950}{0.006} & \est{0.734}{0.009} \\
 & SCRUB & \est{0.016}{0.014} & \est{0.044}{0.027} & \est{0.971}{0.008} & \est{0.749}{0.009} & \est{0.707}{0.017} & \est{0.639}{0.022} & \est{0.961}{0.006} & \est{0.746}{0.008} \\
 & SalUn & \est{0.015}{0.014} & \est{0.034}{0.015} & \est{0.976}{0.003} & \est{0.762}{0.013} & \est{0.687}{0.024} & \est{0.614}{0.031} & \est{0.960}{0.007} & \est{0.744}{0.009} \\
 & SSD & \est{0.039}{0.023} & \est{0.091}{0.025} & \est{0.975}{0.007} & \est{0.764}{0.010} & \est{0.730}{0.017} & \est{0.676}{0.023} & \est{0.975}{0.004} & \est{0.760}{0.011} \\
 & \textbf{ODR} & \textbf{\est{0.023}{0.012}} & \textbf{\est{0.045}{0.015}} & \textbf{\est{0.970}{0.004}} & \textbf{\est{0.754}{0.008}} & \textbf{\est{0.768}{0.014}} & \textbf{\est{0.706}{0.018}} & \textbf{\est{0.983}{0.005}} & \textbf{\est{0.775}{0.011}} \\
\midrule
\multirow{8}{*}{FaceScrub} & Original & \est{0.992}{0.005} & \est{0.975}{0.008} & \est{0.993}{0.003} & \est{0.939}{0.007} & \est{0.992}{0.004} & \est{0.979}{0.005} & \est{0.994}{0.004} & \est{0.942}{0.004} \\
 & Retraining & \est{0.006}{0.006} & \est{0.018}{0.013} & \est{0.991}{0.003} & \est{0.936}{0.005} & \est{0.932}{0.005} & \est{0.904}{0.009} & \est{0.993}{0.003} & \est{0.937}{0.006} \\
 & SISA & \est{0.008}{0.007} & \est{0.022}{0.015} & \est{0.992}{0.002} & \est{0.938}{0.005} & \est{0.934}{0.004} & \est{0.907}{0.007} & \est{0.994}{0.003} & \est{0.939}{0.006} \\
 & Amnesiac & \est{0.061}{0.013} & \est{0.131}{0.010} & \est{0.968}{0.003} & \est{0.904}{0.006} & \est{0.866}{0.012} & \est{0.818}{0.019} & \est{0.974}{0.003} & \est{0.913}{0.005} \\
 & SCRUB & \est{0.034}{0.006} & \est{0.081}{0.013} & \est{0.980}{0.004} & \est{0.916}{0.007} & \est{0.901}{0.006} & \est{0.866}{0.010} & \est{0.983}{0.004} & \est{0.928}{0.005} \\
 & SalUn & \est{0.020}{0.010} & \est{0.056}{0.011} & \est{0.984}{0.002} & \est{0.926}{0.005} & \est{0.893}{0.016} & \est{0.848}{0.019} & \est{0.983}{0.003} & \est{0.925}{0.006} \\
 & SSD & \est{0.086}{0.007} & \est{0.166}{0.016} & \est{0.990}{0.003} & \est{0.936}{0.007} & \est{0.921}{0.009} & \est{0.886}{0.015} & \est{0.990}{0.005} & \est{0.932}{0.004} \\
 & \textbf{ODR} & \textbf{\est{0.013}{0.009}} & \textbf{\est{0.025}{0.012}} & \textbf{\est{0.986}{0.003}} & \textbf{\est{0.929}{0.006}} & \textbf{\est{0.940}{0.006}} & \textbf{\est{0.916}{0.010}} & \textbf{\est{0.996}{0.002}} & \textbf{\est{0.944}{0.004}} \\
\bottomrule
\end{tabular*}
\end{table}

The high deception rates in Table~\ref{tab:odr_dsr} should reproduce the
granularity-specific output behavior expected after unlearning rather than
merely exploit one permissive threshold. We therefore compare ODR with
Retraining, SISA, and the four approximate methods on every dataset and request
level using forgotten-set accuracy $\operatorname{Acc}_u$, forgotten-set
confidence $\operatorname{Conf}_u$, retained-set accuracy
$\operatorname{Acc}_r$, and test accuracy. ODR-Head is evaluated on class
requests, while ODR-Gate replays same-class nonmember posteriors for registered
sample requests. Table~\ref{tab:odr_behavior} shows that ODR-Head reduces
$\operatorname{Acc}_u$ to $0.086$, $0.023$, and $0.013$ on CIFAR10, CIFAR100,
and FaceScrub, respectively, within the range of the evaluated unlearning
methods, while retaining accuracies of $0.982$, $0.970$, and $0.986$. For
sample requests, ODR-Gate produces $\operatorname{Acc}_u$ of $0.936$, $0.768$,
and $0.940$ with confidence of $0.891$, $0.706$, and $0.916$. Its largest
accuracy or confidence gap from Retraining is $0.019$, and retained accuracy
differs from the original model by at most $0.003$. ODR thus reproduces the
distinct class- and sample-level output signatures without sacrificing retained
utility; its deception success cannot be attributed to model collapse or one
isolated observable.

\appsubsection{Construction and Deployment Cost}
\begin{table}[htbp]
\centering
\caption{Construction and deployment cost on CIFAR10 with ResNet18.}
\label{tab:odr_cost}
\fontsize{7.5}{9}\selectfont
\setlength{\tabcolsep}{3pt}
\renewcommand{\arraystretch}{1.12}
\begin{tabular*}{\linewidth}{@{\extracolsep{\fill}}cccccc@{}}
\toprule
Method & Class (s) & Sample (s) & Train$^+$ (MB) & Deploy$^+$ (MB) & Query$^+$ (ms) \\
\midrule
Retraining & $1176.8$ & $1172.4$ & $0$ & $0$ & $0$ \\
SISA & $1126.9$ & $1159.6$ & $1117.5$ & $178.8$ & $3.42$ \\
Amnesiac & $251.8$ & $165.3$ & $1.596{\times}10^{6}$ & $0$ & $0$ \\
SCRUB & $169.4$ & $163.8$ & $44.7$ & $0$ & $0$ \\
SalUn & $88.6$ & $84.9$ & $43.5$ & $0$ & $0$ \\
SSD & $122.0$ & $121.0$ & $89.4$ & $0$ & $0$ \\
\textbf{ODR-Head} & \textbf{$9.3$} & -- & \textbf{$0$} & \textbf{$0$} & \textbf{$0$} \\
\textbf{ODR-Gate} & -- & \textbf{$1.07$} & \textbf{$0$} & \textbf{$0.230$} & \textbf{$0.024$} \\
\bottomrule
\end{tabular*}
\end{table}

To quantify the incentive for output-level deception, we compare its cost with
the evaluated unlearning methods on CIFAR10/ResNet18. Both one forgotten class
and the main $\rho=10\%$ sample request contain 5{,}000 samples. We report class
and sample construction time, method-specific training state, additional
deployed state, and per-query overhead relative to standard single-model
deployment; methods that replace the original model with one model of the same
architecture incur no additional deployment or query cost. The measurements
follow the common protocol and official method mechanisms
\citep{bourtoule2021machine,graves2021amnesiac,kurmanji2023scrub,
fan2024salun,foster2024ssd}. Table~\ref{tab:odr_cost} reports
$1176.8/1172.4$ seconds for class/sample Retraining and
$1126.9$--$1159.6$ seconds for SISA, which additionally stores $1117.5$ MB of
slice checkpoints and deploys $178.8$ MB of shard models. Approximate methods
require $84.9$--$251.8$ seconds; SSD retains its reported ResNet18 means of
$122$ seconds for the class benchmark and $121$ seconds for CIFAR10 random
unlearning with 100 forgotten samples. In contrast, ODR-Head takes $9.3$
seconds, or $0.71\%$ of original training. Including its auxiliary forward
pass, ODR-Gate takes $1.07$ seconds, stores $0.230$ MB of target indices and
replay posteriors, and adds $0.024$ ms per registered query. This cost asymmetry
makes output conformity an economically plausible alternative for a provider
that wishes to retain the original model, explaining why the output evidence
boundary is attractive to manipulate.

\appsection{Supplementary RUV Experiments}
\label{app:ruv_results}

\appsubsection{Decision Margins and Representation Evidence}

\begin{figure}[tbp]
    \centering
    \includegraphics[width=\linewidth]{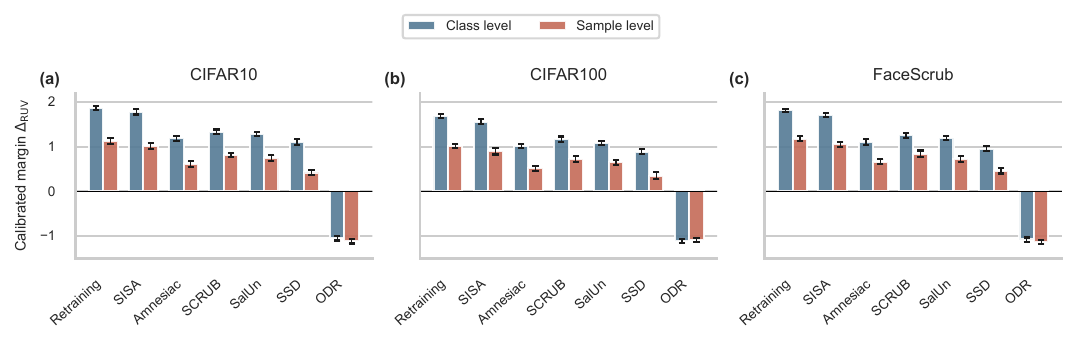}
    \caption{RUV decision margins across datasets and request levels.}
    \label{fig:ruv_evidence}
\end{figure}

To expose how the four representation views produce the decisions summarized
by VA, we record each exact, approximate, and ODR submission's calibrated margin
relative to its request-specific joint permutation boundary and aggregate the
observed standardized ARS, RS, NPG, and MRD effects
$Z_j^{\mathrm{obs}}$ across datasets. Figure~\ref{fig:ruv_evidence} shows
positive mean margins for every evaluated unlearning method and negative
margins for every ODR condition. Across the six dataset-level settings,
unlearning margins range from $0.350$ to $1.872$, whereas ODR margins range from
$-1.188$ to $-1.049$. CIFAR100 sample-level SSD is closest to the boundary,
with margin $0.350$ and pass rate $0.932$, and class margins consistently exceed
sample margins. Table~\ref{tab:ruv_evidence_profile} explains this separation:
Retraining and SISA produce broad four-stage profiles; Amnesiac emphasizes ARS
and RS; SCRUB produces closely matched RS and NPG; SalUn affects activation,
state, and neighborhood evidence; and SSD strengthens from state displacement
toward neighborhood and manifold evidence. ODR induces no target-specific
representation response: all four calibrated effects remain centered near zero
and consequently fall below the positive permutation boundary, producing the
negative margins in Figure~\ref{fig:ruv_evidence}; its largest cross-dataset
mean is $0.020$ at class level and $0.013$ at sample level. The heterogeneous
profiles show why RUV does not require one statistic to dominate every method:
second-largest aggregation accepts a propagation profile only when at least two
stages provide corroborating evidence beyond joint null variation.
\begin{table}[tbp]
\centering
\caption{Cross-dataset calibrated representation evidence profiles ($Z_j^{\mathrm{obs}}$).}
\label{tab:ruv_evidence_profile}
\fontsize{7.5}{9}\selectfont
\setlength{\tabcolsep}{3pt}
\renewcommand{\arraystretch}{1.12}
\begin{tabular*}{\linewidth}{@{\extracolsep{\fill}}ccccc@{}}
\toprule
Submission & ARS & RS & NPG & MRD \\
\midrule
\multicolumn{5}{c}{Class level} \\
\cmidrule(lr){1-5}
Retraining & 2.692 $\pm$ 0.045 & 3.053 $\pm$ 0.072 & 3.320 $\pm$ 0.058 & 3.466 $\pm$ 0.050 \\
SISA & 2.597 $\pm$ 0.045 & 2.987 $\pm$ 0.068 & 3.271 $\pm$ 0.065 & 3.410 $\pm$ 0.073 \\
Amnesiac & 2.909 $\pm$ 0.181 & 2.674 $\pm$ 0.110 & 2.628 $\pm$ 0.042 & 2.373 $\pm$ 0.116 \\
SCRUB & 2.259 $\pm$ 0.083 & 2.847 $\pm$ 0.088 & 2.821 $\pm$ 0.194 & 2.622 $\pm$ 0.116 \\
SalUn & 2.563 $\pm$ 0.111 & 2.529 $\pm$ 0.081 & 2.832 $\pm$ 0.094 & 2.682 $\pm$ 0.043 \\
SSD & 1.677 $\pm$ 0.257 & 2.170 $\pm$ 0.182 & 2.577 $\pm$ 0.102 & 2.846 $\pm$ 0.097 \\
ODR & 0.013 $\pm$ 0.045 & 0.010 $\pm$ 0.046 & 0.017 $\pm$ 0.065 & 0.020 $\pm$ 0.056 \\
\midrule
\multicolumn{5}{c}{Sample level} \\
\cmidrule(lr){1-5}
Retraining & 2.147 $\pm$ 0.117 & 2.031 $\pm$ 0.115 & 2.174 $\pm$ 0.130 & 1.973 $\pm$ 0.084 \\
SISA & 2.093 $\pm$ 0.120 & 1.948 $\pm$ 0.117 & 2.113 $\pm$ 0.149 & 1.914 $\pm$ 0.122 \\
Amnesiac & 2.105 $\pm$ 0.158 & 2.029 $\pm$ 0.168 & 1.922 $\pm$ 0.110 & 1.716 $\pm$ 0.180 \\
SCRUB & 1.639 $\pm$ 0.126 & 2.041 $\pm$ 0.091 & 2.046 $\pm$ 0.194 & 1.745 $\pm$ 0.173 \\
SalUn & 1.811 $\pm$ 0.158 & 1.764 $\pm$ 0.101 & 2.065 $\pm$ 0.173 & 1.860 $\pm$ 0.137 \\
SSD & 0.955 $\pm$ 0.145 & 1.333 $\pm$ 0.107 & 1.752 $\pm$ 0.159 & 1.975 $\pm$ 0.124 \\
ODR & -0.003 $\pm$ 0.032 & -0.000 $\pm$ 0.044 & 0.013 $\pm$ 0.045 & 0.007 $\pm$ 0.031 \\
\bottomrule
\end{tabular*}
\end{table}

\appsubsection{Verification Overhead}

\begin{table}[htbp]
\centering
\caption{Verification overhead on CIFAR10 with ResNet18.}
\label{tab:verification_overhead}
\fontsize{7.5}{9}\selectfont
\setlength{\tabcolsep}{3pt}
\renewcommand{\arraystretch}{1.12}
\begin{tabular*}{\linewidth}{@{\extracolsep{\fill}}cccccc@{}}
\toprule
Method & Requirement & Verify t (s) & Extra t (s) & Verify mem. & Extra mem. \\
\midrule
Accuracy & Output inference & 0.84 & 0 & 812 & 0 \\
LiRA-16 & 16 shadow models & 1.62 & 20918.4 & 1136 & 715.2 \\
TAPE & Shadows + AE & 1.13 & 113.0 & 886 & 144.0 \\
TruVRF & 2 sensitivity passes & 91.20 & 0 & 2528 & 0 \\
RUV & Multilayer forward & 4.37 & 0 & 1684 & 0 \\
\bottomrule
\end{tabular*}
\end{table}

We measure the computation and memory required by forward representation
auditing relative to output and gradient verification on CIFAR10/ResNet18 with
$\rho=10\%$ sample requests and batch size 128. Online time and peak GPU memory
cover one audit after required assets are loaded, while extra time and memory
cover offline preparation and persistent auxiliary states. LiRA-16 uses 16
shadow models and the global variance estimator \citep{carlini2022lira}; TAPE
includes influence-based shadow construction and reconstructor training; and
TruVRF follows Metric-III with two model-sensitivity extractions
\citep{zhou2025truvrf}. Each online measurement is repeated ten times on an
NVIDIA RTX 4090 GPU and Intel Xeon Bronze 3204C CPU.
Table~\ref{tab:verification_overhead} reports $0.84$ seconds and $812$ MB for
Accuracy. LiRA-16 and TAPE take $1.62$ and $1.13$ seconds online but require
$20{,}918.4$ and $113.0$ seconds of preparation, storing $715.2$ and $144.0$ MB
of auxiliary state. TruVRF's two backward sensitivity extractions take $91.2$
seconds and peak at $2528$ MB. RUV requires no separately trained verifier and
completes in $4.37$ seconds with a peak of $1684$ MB. It is therefore more
involved than a direct output check, as expected for multiple representation
blocks, but avoids both shadow-model preparation and backward computation,
reducing online time by $95.2\%$ and peak GPU memory by $33.4\%$ relative to
TruVRF while retaining an internal evidence boundary.

\appsubsection{Representation Selection}
\label{app:ablation}

The following ablations use the same controlled set of 100 class-level and 100
sample-level audit instances, comprising ten requests for each of ten original
model seeds. Every variant reuses these instances, so its standard RUV entry is
a matched baseline rather than a reaggregation of the dataset-specific results
in Table~\ref{tab:verification_main}.

\begin{table}[htbp]
\centering
\caption{Verification accuracy under representation-selection strategies.}
\label{tab:representation_selection}
\fontsize{7.5}{9}\selectfont
\setlength{\tabcolsep}{3pt}
\renewcommand{\arraystretch}{1.12}
\begin{tabular*}{0.78\linewidth}{@{\extracolsep{\fill}}ccc@{}}
\toprule
\multicolumn{1}{c}{Selection strategy} & Class level & Sample level \\
\midrule
Preclassifier only & \textbf{0.991 $\pm$ 0.006} & 0.742 $\pm$ 0.022 \\
All nonterminal & 0.901 $\pm$ 0.015 & \textbf{0.957 $\pm$ 0.008} \\
All representations & 0.944 $\pm$ 0.012 & 0.913 $\pm$ 0.015 \\
\bottomrule
\end{tabular*}
\end{table}

To identify the representation blocks appropriate to each request level, we
compare three architecture-independent strategies under otherwise identical
RUV settings: the preclassifier representation alone, all nonterminal blocks,
and all available representations. Table~\ref{tab:representation_selection}
shows that the preclassifier block gives the strongest class result
($0.991\pm0.006$) but reduces sample VA to $0.742\pm0.022$. Aggregating all
nonterminal blocks reverses the pattern and yields the strongest sample result
($0.957\pm0.008$), whereas using all representations dilutes the
granularity-specific signal at both levels. This reversal supports the rule in
Eq.~\eqref{eq:main_layers}: terminal representations are organized around class
means coupled to the classifier \citep{yosinski2014transferable,
papyan2020prevalence}, while instance-specific memorization can be distributed
across nonterminal blocks \citep{maini2023localized}.

\appsubsection{Joint Calibration}

\begin{table}[htbp]
\centering
\caption{Joint calibration and decision-rule ablation.}
\label{tab:calibration_ablation}
\fontsize{7.5}{9}\selectfont
\setlength{\tabcolsep}{3pt}
\renewcommand{\arraystretch}{1.12}
\begin{tabular*}{0.78\linewidth}{@{\extracolsep{\fill}}ccc@{}}
\toprule
\multicolumn{1}{c}{Decision rule} & Class level & Sample level \\
\midrule
\textbf{Full RUV} & \textbf{0.991 $\pm$ 0.006} & \textbf{0.957 $\pm$ 0.008} \\
Independent permutation & 0.964 $\pm$ 0.012 & 0.925 $\pm$ 0.015 \\
Fixed global threshold & 0.906 $\pm$ 0.022 & 0.848 $\pm$ 0.020 \\
\bottomrule
\end{tabular*}
\end{table}

To isolate the contribution of request-specific joint calibration, we compare
full RUV with two alternatives while leaving the evidence views and
target-specific gaps unchanged. Independent permutation destroys cross-view
dependence before aggregation, while a fixed threshold replaces the
request-specific null boundary with one global decision boundary.
Table~\ref{tab:calibration_ablation} shows that independent permutation reduces
class/sample VA from $0.991/0.957$ to $0.964/0.925$, and the fixed threshold
causes a larger reduction to $0.906/0.848$. Joint permutation therefore
preserves the empirical dependence among ARS, RS, NPG, and MRD, while
request-specific calibration adapts the boundary to the current model pair,
request, and audit sample. Together they avoid four incompatible marginal tests
and one brittle global constant.

\begin{figure}[h]
    \centering
    \includegraphics[width=0.68\linewidth]{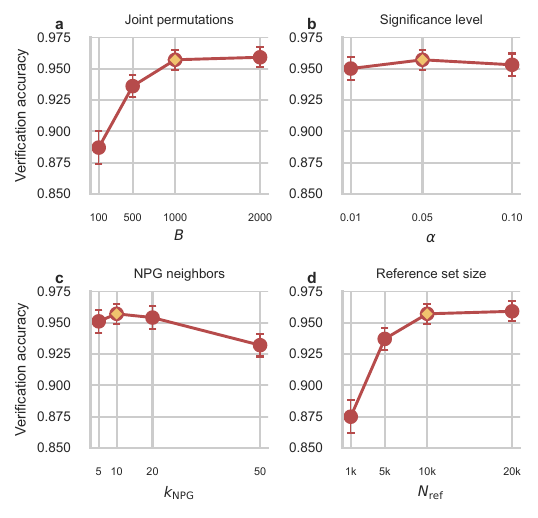}
    \caption{Sample-level RUV parameter sensitivity.}
    \label{fig:parameter_sensitivity}
\end{figure}

\appsubsection{Parameter Sensitivity}

We test whether sample-level performance depends on a narrow hyperparameter
configuration by varying one factor at a time over permutation count $B$, test
level $\alpha$, neighborhood size $k_{\mathrm{NPG}}$, and reference-set size
$N_{\mathrm{ref}}$, while fixing all remaining settings and audit instances;
diamonds in Figure~\ref{fig:parameter_sensitivity} mark the defaults. Accuracy
rises from $0.887$ at $B=100$ to $0.957$ at $B=1000$, while doubling the
permutations adds only $0.002$. It remains between $0.950$ and $0.957$ for
$\alpha\in\{0.01,0.05,0.10\}$. NPG is stable for
$k_{\mathrm{NPG}}\in\{5,10,20\}$ and declines at $50$, where increasingly
global relations weaken local evidence. Increasing $N_{\mathrm{ref}}$ from
$1{,}000$ to $10{,}000$ raises VA from $0.875$ to $0.957$, while $20{,}000$
provides no material gain. The defaults $B=1000$, $\alpha=0.05$,
$k_{\mathrm{NPG}}=10$, and $N_{\mathrm{ref}}=10{,}000$ consequently lie on
broad performance plateaus that provide sufficient statistical and geometric
resolution without unnecessary audit cost.
\WFclear

\appsubsection{Request-Scale Robustness}
\label{sec:request_scale_robustness}

\begin{figure}
    \centering
    \includegraphics[width=0.68\linewidth]{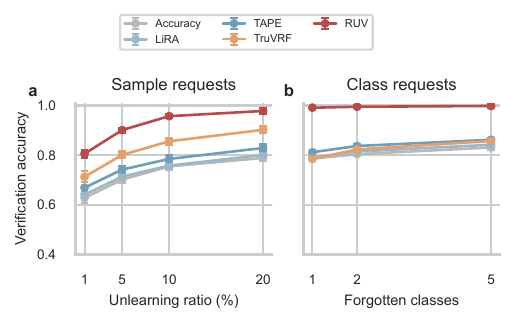}
    \caption{Verification accuracy across request scales.}
    \label{fig:request_scale}
\end{figure}

To examine how verification changes with the amount of target evidence, we vary
the CIFAR10/ResNet18 sample request ratio over
$\{1\%,5\%,10\%,20\%\}$ and the class request over $\{1,2,5\}$ forgotten
classes. The original-model family, request manifests at each scale, compared
verifiers, representation-selection rule, retained-control matching, reference
size, and joint calibration remain fixed. Figure~\ref{fig:request_scale} shows
that sample requests are most challenging at $1\%$: RUV reaches $0.806$,
compared with $0.713$ for TruVRF and $0.669$ for TAPE. RUV increases to $0.957$
at $10\%$ and $0.978$ at $20\%$, remaining $7.6$--$10.0$ percentage points
above TruVRF. Class evidence is stronger, increasing from $0.991$ for one
forgotten class to $0.998$ for five and exceeding the strongest competitor by
$13.6$--$17.9$ points. Verification thus improves as requests provide more
target evidence, while RUV remains strongest at every scale; the contrasting
curves also show why coherent class removal is visible at smaller scales than
distributed sample removal.

\end{document}